\documentclass[sigconf]{acmart}
\AtBeginDocument{%
  }

\copyrightyear{2018}
\acmYear{2018}
\setcopyright{acmlicensed}
\acmConference[Conference acronym 'XX]{Make sure to enter the correct
  conference title from your rights confirmation email}{June 03--05,
  2018}{Woodstock, NY}
\acmDOI{XXXXXXX.XXXXXXX}

\acmISBN{978-1-4503-XXXX-X/2018/06}

\usepackage{setspace}
\usepackage{xcolor}
\usepackage{xspace}
\usepackage{pifont}
\usepackage{multirow}
\usepackage{lipsum}
\usepackage{placeins}
\usepackage{enumitem}
\usepackage{hhline}      
\usepackage{float}
\usepackage[
  group-separator={,},  
  group-minimum-digits=4  
]{siunitx}

\usepackage{siunitx}
\begin{document}

\newcommand{\xmark}{\ding{55}}
\newcommand\zty[1]{{\color{black}#1}}
\newcommand{\name}[0]{\textbf{M}$^3$\textbf{ob}\xspace}

\newlist{parenum}{enumerate}{1}
\setlist[parenum,1]{label=\arabic*), ref=\arabic*)}

\title{Learning Multi-Modal Mobility Dynamics for Generalized Next Location Recommendation}
\author{Junshu Dai}
\affiliation{
  \institution{Zhejiang University}
  \city{Hangzhou}
  \state{Zhejiang}
  \country{China}
}
\email{djs@zju.edu.cn}

\author{Yu Wang}
\affiliation{
  \institution{Zhejiang University}
  \city{Hangzhou}
  \state{Zhejiang}
  \country{China}
}
\email{yu.wang@zju.edu.cn}

\author{Tongya Zheng}
\authornote{Corresponding author.}
\affiliation{
  \institution{Hangzhou City University}
  \city{Hangzhou}
  \state{Zhejiang}
  \country{China}
}
\email{doujiang_zheng@163.com	}

\author{Wei Ji}
\affiliation{
  \institution{Nanjing University}
  \city{Nanjing}
  \state{Jiangsu}
  \country{China}
}
\email{weiji0523@gmail.com}

\author{Qinghong Guo}
\affiliation{
  \institution{Zhejiang University}
  \city{Hangzhou}
  \state{Zhejiang}
  \country{China}
}
\email{q.h_guo@zju.edu.cn}

\author{Ji Cao}
\affiliation{
  \institution{Zhejiang University}
  \city{Hangzhou}
  \state{Zhejiang}
  \country{China}
}
\email{caoj25@zju.edu.cn}

\author{Jie Song}
\affiliation{
  \institution{Zhejiang University}
  \city{Hangzhou}
  \state{Zhejiang}
  \country{China}
}
\email{sjie@zju.edu.cn}

\author{Canghong Jin}
\affiliation{
  \institution{Hangzhou City University}
  \city{Hangzhou}
  \state{Zhejiang}
  \country{China}
}
\email{jinch@zucc.edu.cn}

\author{Mingli Song}
\affiliation{
  \institution{Zhejiang University}
  \city{Hangzhou}
  \state{Zhejiang}
  \country{China}
}
\email{brooksong@zju.edu.cn}


\renewcommand{\shortauthors}{Dai et al.}

\begin{abstract}
\begin{spacing}{0.95}
  The precise prediction of human mobility has produced significant socioeconomic impacts, such as location recommendations and evacuation suggestions. However, existing methods suffer from limited generalization capability: unimodal approaches are constrained by data sparsity and inherent biases, while multi-modal methods struggle to effectively capture mobility dynamics caused by the semantic gap between static multi-modal representation and spatial-temporal dynamics. Therefore, we leverage multi-modal spatial-temporal knowledge to characterize mobility dynamics for the location recommendation task, dubbed as \textbf{M}ulti-\textbf{M}odal \textbf{Mob}ility (\textbf{M}$^3$\textbf{ob}). 
  First, we construct a unified spatial-temporal relational graph (STRG) for multi-modal representation, by leveraging the functional semantics and spatial-temporal knowledge captured by the large language models (LLMs)-enhanced spatial-temporal knowledge graph (STKG).
  Second, we design a gating mechanism to fuse spatial-temporal graph representations of different modalities, and propose an STKG-guided cross-modal alignment to inject spatial-temporal dynamic knowledge into the static image modality.
  Extensive experiments on six public datasets show that our proposed method not only achieves consistent improvements in normal scenarios but also exhibits significant generalization ability in abnormal scenarios. 
  Our code will be made publicly available at \url{https://anonymous.4open.science/r/M3ob-62EF}.
\end{spacing}
  
\end{abstract}




\maketitle

\section{Introduction}

The widely adopted mobile devices have made location-based services highly accessible on the Web, allowing applications such as location search~\cite{lane2010hapori,yang2013fine}, mobile navigation~\cite{wu2011location}, and location recommendation~\cite{zhang2014lore,liu2016exploring}. However, while the related check-in sequence data greatly facilitates the integration of web services into daily life, two issues lead to data sparsity: the sheer number of locations, and users’ tendency to visit only a limited number of popular locations of interest. Furthermore, abnormal mobility behaviors are difficult to capture due to their scarce occurrence, such as visits to long-tail locations or during extreme weather events.

The remarkable success of deep learning has transformed next location recommendation into a representation learning paradigm, which can be categorized into sequence-based methods and graph-based methods.
Sequence-based methods design spatial-temporal modules to extract mobility representations within historical check-in sequences, including markov chain-based model~\cite{gambs2012next}, RNNs-based model~\cite{sun2020go,yang2020location} and attention-based model~\cite{feng2018deepmove,luo2021stan,duan2023clsprec,sun2024going}. 
Graph-based methods~\cite{han2020stgcn,yang2022getnext,lim2022hierarchical,rao2022graph,chen2022building,qin2023diffusion,wang2023adaptive,qin2023disenpoi,yan2023spatio,rao2024next} enhance location representations by sharing common patterns of locations based on spatial-temporal graphs.
However, the sparsity of check-in sequences and the inherent mobility bias lead to suboptimal performance in representation learning for these unimodal approaches.



Recently, language and vision models have demonstrated advanced representations based on sufficient self-supervised learning, which inspires the progress of multi-modal learning across various domains.
In e-commerce recommendation, MKGAT~\cite{sun2020multi} and MMGCN~\cite{wei2019mmgcn} leverage multi-modal attributes of items together with abundant user-item interactions to construct a multi-modal graph for learning. Alignrec~\cite{liu2024alignrec} narrows the representation distance through contrastive learning between items and their multi-modal content.
In location recommendation, to our knowledge, MMPOI~\cite{xu2024mmpoi} was the first to introduce multi-modality by constructing a multi-modal graph via cosine similarity. However, existing multi-modal approaches neglect the semantic gap between mobility dynamics and static multi-modal representations, thus inevitably leading to performance degradation when generalizing.

Despite the effectiveness of multi-modal pre-training representations, bridging the semantic heterogeneity between static multi-modal data and dynamic human mobility remains a key challenge.
\textbf{(1) How to model the dynamic relationships of locations and users in a multi-modal view?}
Current multi-modal methods construct static multi-modal relationships to obtain multi-modal knowledge without integrated mobility dynamics, leading to limited generalization capability in dynamic scenarios.
\textbf{(2) How to align multi-modal representations of locations and users to capture mobility dynamics?}
The inherent semantic gap between different modalities necessitates spatial-temporal-aware alignment techniques to eliminate the modality heterogeneity.

To this end, we design a \textbf{M}ulti-\textbf{M}odal \textbf{Mob}ility (\name) framework that overcomes the generalization gap by building a shared graph across multiple modalities. 

To address the first challenge of modeling multi-modal dynamics, we build a spatial-temporal relational graph (STRG) to enable the sharing of spatial-temporal knowledge across modalities, which leverages the functional semantics and spatial-temporal knowledge from a large language models (LLMs)-enhanced spatial-temporal knowledge graph (STKG). The hierarchical knowledge from the STKG further enables the modeling of multi-level user preferences.
For the second challenge of aligning multi-modal representations, we employ a gating mechanism to dynamically fuse spatial-temporal graph representations from different modalities, thereby mitigating modal interference. Additionally, STKG-guided cross-modal alignment is applied to reduce inter-modal discrepancies.

Our main contributions are summarized as follows:
\begin{itemize}[leftmargin=*]
\item 
We propose \name, a novel framework for seamlessly fusing and aligning dynamic human mobility patterns with static multi-modal representations.
\item 
We construct a new STRG based on LLMs-enhanced STKG to unify multi-modal mobility representations and leverage STKG-guided cross-modal alignment to improve the spatial-temporal dynamics of static image modality.
\item 
Experimental results show that \name achieves a superior performance gain over the baseline, with robust generalization in adverse weather and long-tail scenarios, and high efficiency.
\end{itemize}
\vspace{-11pt}

\section{Related Work}


\subsection{Next Location Recommendation}
Deep learning approaches have dominated the advanced progress of next location recommendation, which predicts users' future visits and can be categorized into sequence-based and graph-based models. 
\textbf{Sequence-based models} capture spatial-temporal patterns through Markov chains~\cite{gambs2012next}, RNN architectures~\cite{feng2018deepmove,sun2020go,yang2020location,zhao2020go}, or attention mechanisms~\cite{luo2021stan,sun2024going}. 
For instance, LSTPM~\cite{sun2020go} employs a non-local network and a geographically dilated RNN to model long- and short-term preferences. 
CLSPRec~\cite{duan2023clsprec} leverages contrastive learning on raw sequential data to effectively distinguish between long- and short-term user preferences. MCLP~\cite{sun2024going} incorporates a multi-head attention mechanism to generate arrival time embeddings as contextual information for location recommendation. However, these methods insufficiently capture the spatial-temporal transition patterns of different locations in trajectories.
\textbf{Graph-based models} primarily utilize spatial-temporal relationships between locations and apply GNNs to enhance location representations. A line of works enhance location embeddings by constructing spatial-temporal graphs based on geographical and transitional relationships between locations~\cite{han2020stgcn,yang2022getnext,lim2022hierarchical,qin2023diffusion}. GETNEXT~\cite{yang2022getnext} constructs a global trajectory flow graph to better capture transition patterns between locations. Another line of works builds spatial-temporal mobility graphs using pre-trained spatial-temporal knowledge graphs~\cite{rao2022graph,yin2023next,xu2024taming}. 
AGCL~\cite{rao2024next} employs graph-enhanced location representation and negative-sample contrast to enhance the discriminability and consistency of POI representations. LoTNext~\cite{xu2024taming} introduces long-tailed adjustment strategies for both graph and loss to tackle the imbalance between head and tail locations. However, unimodal check-in sequences inevitably limit the representation quality of these methods, yielding  suboptimal generalization capability in scarce mobility scenarios.
Recent studies have attempted to alleviate this sparsity issue by enriching the semantic features of locations with multi-modal information. MMPOI~\cite{xu2024mmpoi} constructs static multi-modal graphs using intra-modal cosine similarity to leverage independent information from each modality. TSPN-RA~\cite{jiang2024towards} builds a spatial graph based on a static quadtree structure derived from remote sensing imagery to incorporate real-world environmental semantics. However, static relationships of locations extracted from multi-modal mobility data overlook the spatial-temporal dynamics, thereby leaving significant semantic gaps between multi-modal embedding and mobility learning. Table~\ref{tab:novel} illustrates that we inject mobility dynamics into multi-modal representation beyond limitations of existing methods.
\begin{table}[t]
    \small
    \caption{Comparison between our \name against baseline methods. STKG denotes Spatial-Temporal Knowledge Graph.}
  \begin{tabular}{l|ccc}
    \hline
    \multirow{2}{*}{\textbf{Methods}} &\multicolumn{2}{c}{\textbf{Location}}& \textbf{User} \\ 
    &\textbf{Multi-modal Graph} &\textbf{Alignment}&\textbf{Preference}\\
    
    \hline
    \textbf{GETNEXT}~\cite{yang2022getnext} & \xmark & \xmark & Single-level\\
    \textbf{DisenPOI}~\cite{qin2023disenpoi} & \xmark & Data-based & Single-level\\
    \textbf{Diff-POI}~\cite{qin2023diffusion}& \xmark& \xmark& Single-level\\
    \textbf{STHGCN}~\cite{yan2023spatio}& \xmark& \xmark& Single-level \\
    \textbf{CLSPRec}~\cite{duan2023clsprec}& \xmark& Data-based& Single-level\\
    \textbf{AGCL}~\cite{rao2024next}& \xmark& Graph-based & Single-level\\
    \textbf{MCLP}~\cite{sun2024going}& \xmark& \xmark& Single-level\\
    \textbf{LoTNext}~\cite{xu2024taming}& \xmark& \xmark& Single-level\\
    \textbf{MMPOI}~\cite{xu2024mmpoi}& Cosine Similarity& \xmark& Single-level\\
    \textbf{TSPN-RA}~\cite{jiang2024towards}& Quad Tree& \xmark& Single-level\\
    \name\textbf{(Ours)}& STKG-based& STKG-based & Multi-level\\
  \hline
\end{tabular}

\label{tab:novel}
\end{table}

\subsection{Multi-Modal Recommendation}
Multi-modal content, which enriches item information and captures users’ fine-grained preferences, has been widely introduced into recommendation systems in recent studies~\cite{zhou2023comprehensive,liu2024multimodal}. Existing research has effectively integrated multi-modal information through the integration of attention mechanisms~\cite{liu2019user,liu2019nrpa,chen2021cmbf,liu2021noninvasive}. Another line of work constructs multi-modal graphs based on user–item interactions and item multi-modal attributes to enrich item semantic representations~\cite{wei2019mmgcn,wei2020graph,tao2020mgat,wang2021dualgnn}. Further advancing this direction, several studies leverage multi-modal knowledge graphs to model complex interactions between users and multi-modal information~\cite{sun2020multi,wang2020enhanced,liu2022multi}. 
Another approach~\cite{zhou2023bootstrap,liu2024alignrec} seeks alignment in the representation space, diminishing the distance between representations through the adoption of contrastive learning between items and their multi-modal content.
Existing multi-modal recommendation primarily focuses on static modal fusion and fails to model the inherent spatiotemporal dynamics of mobile data, making it difficult to be directly applied to mobility representation tasks.


\section{Preliminary}
\textbf{Trajectory Sequence.}
The set of locations is denoted as \(P=\{p_1,p_2,\ldots,p_{N_{p}}\}\) and the set of users is $ U = \{u_1,u_2,\ldots,u_{N_u}\}$, where $N_p$ and $N_u$ are the total number of locations and users respectively. 
Each record is denoted by a tuple $q=\{u,p,t\}$, indicating that user $u$ visited location $p$ at time $t$.
The trajectory of each user $u \in U $ is denoted by $ S_u=\{q_u^1,q_u^2,\ldots\}$.\\
\textbf{Multi-modal Location.}
A location $p$ can be represented in either an ID modality or a pair of latitude and longitude.
In this paper, we augment the location semantics by textual and visual modalities.
Moreover, the textual modality of a location is extended to hierarchical semantics, from the location category $c$ to the user activity $a$, denoted by $p_\textrm{text}=\{p,c,a\}$.
On the other hand, the visual modality of a location is denoted by remote sensing images at different zoom levels, denoted by $p_\textrm{img}=\{p_\textrm{img}^h | h \in \mathcal{H}\}$, where $\mathcal{H} = \{\text{coarse}, \text{medium}, \text{fine}\}$.\\
\textbf{Next Location Recommendation.}
Given a user's trajectory sequence $ \{q_u^1,q_u^2,\ldots,q_u^n\}$, our goal is to predict the next location $ p_{n+1} $ that user $u$ most likely to visit.

\section{Methodology}

\begin{figure*}[t]
	\centering
	\includegraphics[width=\textwidth]{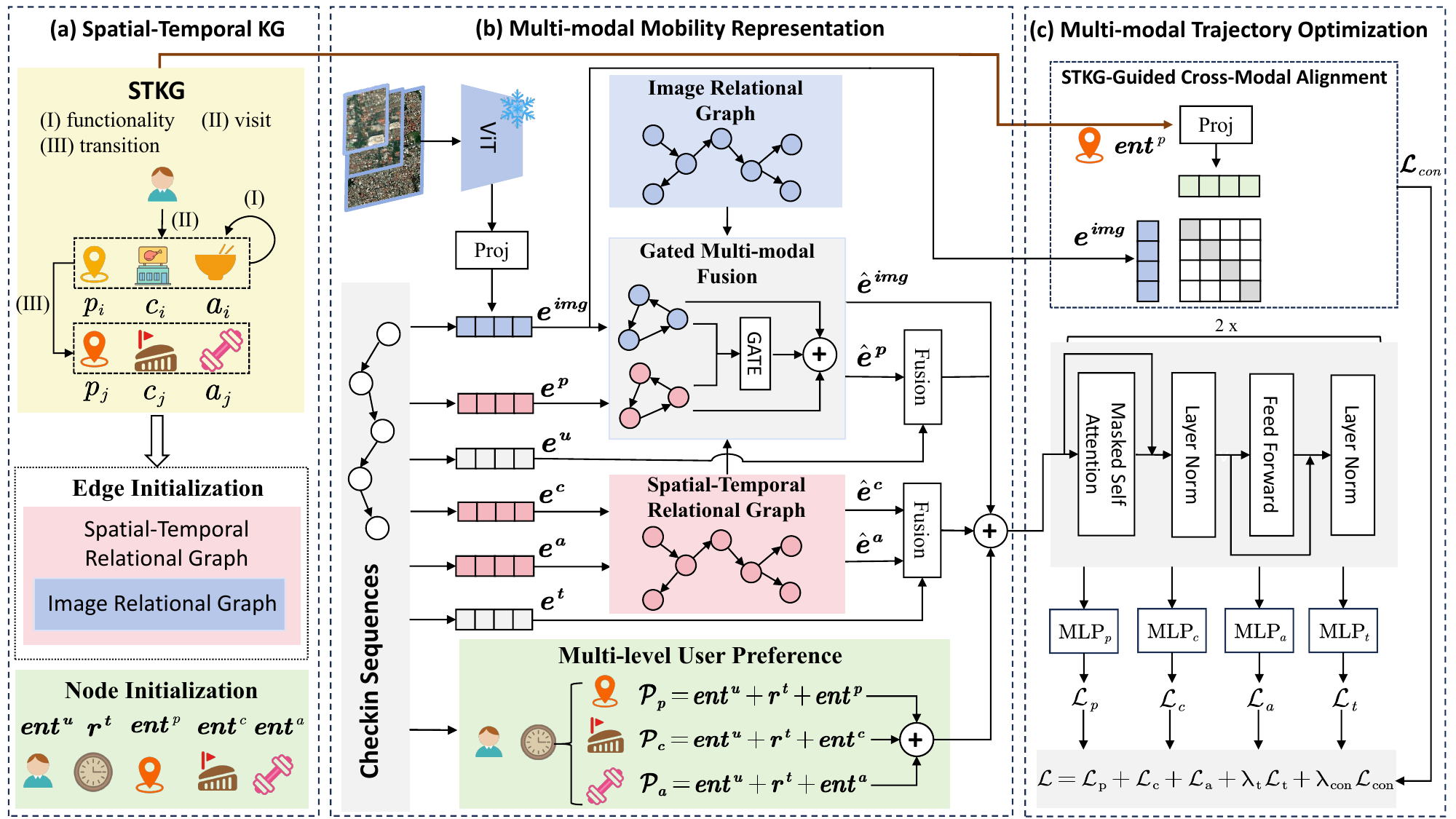} %
	\caption{The illustration diagram of our proposed \textbf{M}ulti-\textbf{M}odal \textbf{Mob}ility representation (\name) framework. Among these, spatial-temporal knowledge graph, spatial-temporal relational graph, and image relational graph are responsible for modeling multi-modal dynamics, while gated multi-modal fusion and STKG-guided cross-modal alignment enable multi-modal alignment.}
\label{eq:framework}
\end{figure*}

Figure~\ref{eq:framework} demonstrates that the overall workflow of our proposed \name framework can be divided into three parts.
(a) Spatial-Temporal Knowledge Graph (STKG) enriches the spatial-temporal semantics by LLM-enhanced textual semantics of spatial-temporal entities.
(b) Leveraging the unified embedding space of the Spatial-Temporal Knowledge Graph (STKG), the Multi-modal Mobility Representation first constructs a novel Spatial-Temporal Relational Graph (STRG). This graph is built using spatial-temporal dynamics embedded in STKG-fused mobility dynamics, and it supports multi-modal location fusion. Additionally, the Multi-modal Mobility Representation builds multi-level user preferences by leveraging textual semantics that range from coarse to fine granularity.
(c) Multi-modal Trajectory Optimization is regularized by hierarchical semantic prediction and STKG-guided cross-modal alignment to incorporate multi-modal spatial-temporal knowledge.
\subsection{Spatial-Temporal Knowledge Graph}
High-order textual semantics alleviates long-tail location data sparsity and enrich the entity association dimensions of SpatioTemporal Knowledge Graph (STKG). Distinct from the existing works~\cite{rao2022graph,xu2024taming}, we are the first to inject textual semantics derived from Large Language Models (LLMs) into STKG, thereby obtaining hierarchical spatiotemporal dynamic relationships.
We first applied DeepSeek-R1\footnote{https://www.deepseek.com/} to cluster location-based categorical texts, thereby obtaining 12 general human activity types. This process resulted in a three-level semantic hierarchy, ranging from coarse to fine: activity, category, and location. The prompt for activity generation is provided in Appendix A.1. This LLM-based approach not only enhances mapping accuracy but also addresses the challenges arising from divergent category definitions across different urban contexts.

Formally, a Hierarchical STKG is defined as a multi-relation graph $\mathcal{G}=(\mathcal{E},\mathcal{R},\mathcal{F})$, where $\mathcal{E}$ is the set of entities, $\mathcal{R}$ is the set of relations and $\mathcal{F}=\{(e^\textrm{head},r,e^\textrm{tail})\}$, where $\textrm{head},\textrm{tail}\in\mathcal{E},r\in\mathcal{R}$.
Each triplet $(e^\textrm{head},r,e^\textrm{tail})$ denotes an edge in KG from head to tail with relation $r$.
We consider users, locations, and their associated categories and activities as spatial-temporal entities to enhance knowledge capacity and characterize relations of entities as follows.

\begin{itemize}[leftmargin=*]
 \item 
 \textbf{Functionality Knowledge}. Hierarchical relations $r_f$ describe the mappings among multi-level entities (location, category, and activity): $(p_i, r_f, c_j)$ and $(c_j, r_f, a_k)$.
\item \textbf{Visit relation of Mobility Knowledge}: Visit relation $r^t$ is defined as a user's interaction with a location, along with its associated categories and activities, represented by the triplets $(u, r^t, p)$, $(u, r^t, c)$, and $(u, r^t, a)$. Here, the temporal context $r^t$ is discretized into 48 time slots.
\item \textbf{Transition relation of Mobility Knowledge}: Transition relation $r_t$ models entity transitions in trajectories between locations $(p_i, r_t, p_j)$, categories $(c_i, r_t, c_j)$, and activities $(a_i, r_t, a_j)$, capturing sequential behavior.

\end{itemize}

We pretrain STKG using TransE~\cite{bordes2013translating} and the resulting embeddings are then utilized in subsequent modules to construct the spatial-temporal relation graph, which is further adopted to fuse the multi-modal representations of locations.

\subsection{Multi-modal Mobility Representation}

\subsubsection{Spatial-Temporal Relational Graph.}
We construct a data-driven Spatial-Temporal Relational Graph by leveraging entity similarity in the STKG, which establishes unified dynamic relationships for multi-modal data and effectively mitigates the issue of data sparsity.
The similarity function of multi-level entities is computed by
\begin{equation}
\label{eq:similarity}
\begin{split}
    \textrm{sim}(e_i,e_j) &= \exp\big(-d(e_i,e_j)\big), e\in{\{p,c,a\}},\\
    d(e_i,e_j) &= \lVert \textrm{ent}^{e_i}+r_t-\textrm{ent}^{e_j}\rVert, e\in{\{p,c,a\}},\\
\end{split}
\end{equation}
where $\textrm{ent}^{k_i}, \textrm{ent}^{k_j}, r_t$ are obtained from the pretrained STKG. Using the hierarchical transition edge weights computed via Equation~\ref{eq:similarity}, we establish three mobility transition matrices $M^p \in N_p \times N_p$, $M^c \in N_c \times N_c$, $M^a \in N_a \times N_a$. $N_c$ and $N_a$ are the number of categories and activities, respectively.

We further reduce useless graph edges by selecting from $k$ nearest neighbors ($k$-NN), written as:
\begin{equation}
\begin{split}
G_{i,j}^e &=
\begin{cases}
M_{i,j}^e, & \text{if Entity } e_j \in \mathcal{N}_k(e_i) \\
 0, & \text{Otherwise}
\end{cases}
\end{split}
,e \in \{p,c,a\},
\end{equation}
where $\mathcal{N}_k(e_i)$ is the $k$ nearest neighbors of different entities.
The STRG is further normalized by dividing a diagonal matrix $D_e$ of row-wise maximum values.
The corresponding representations of different entities are obtained by GCN, written as:
\begin{equation}
\mathbf{Z}^\textrm{fusion}_e = \sigma(D_e^\text{-1} G^e \mathbf{Z}^e W) ,e\in{\{p,c,a\}},
\end{equation}
where $\mathbf{Z}^e$ denotes the original one-hot embeddings of a type of entity, $G^e$ is the corresponding STRG, $W$ is a weight matrix, and $\sigma(\cdot)$ is the activation function.

Furthermore, we update the representations by integrating the spatial-temporal graph representations of both categories and activities with their original features through residual connections.
\begin{equation}
\hat{\mathbf{Z}}^e = \mathbf{Z}^\textrm{fusion}_e+\mathbf{Z}^e,e\in{\{c,a\}},
\end{equation}

\subsubsection{Multi-scale Image Representation.}
In the real world, a direct correspondence exists between a remote sensing image and its geographic location, as such images capture the surrounding environmental semantics. Remote sensing images exhibit hierarchical semantics: lower zoom levels capture broader regional context, while higher levels provide finer-grained details. To effectively leverage multi-scale representations, we first extract features from localized images at three zoom levels using pre-trained CLIP vision encoder (ViT). These features are then projected into the ID modality's semantic space via modality-specific MLP layers ($\text{Proj}_h$):
\begin{equation}
\begin{split}
    \mathbf{Z}^\textrm{img} &= 
    \sum_{h\in \mathcal{H}}
    \textrm{Proj}_h(\textrm{Image-Encoder}(p_\textrm{img}^h)).
\end{split}
\end{equation}
To prevent catastrophic forgetting and reduce GPU memory requirements, we freeze the parameters of all pretrained image encoders during the following training tasks.\\
\textbf{Image Relational Graph.} We introduce mobility dynamic relationships into the remote sensing imagery to overcome the limitation of static graphs based on cosine similarity between images of previous methods~\cite{xu2024mmpoi}. To achieve this, we share the Spatial-Temporal-Relational Graph (STRG) structure of a location with its corresponding image to build an image relational graph $G^\textrm{img}$. The corresponding representations of image modality are obtained by GCN, written as:
\begin{equation}
    \mathbf{Z}^\textrm{fusion}_\textrm{img} = \sigma(D_\textrm{img}^\text{-1} G^\textrm{img}\mathbf{Z}^\textrm{img}W),
\end{equation} 
where $D_\textrm{img}$ also denotes the row-wise max diagonal matrix of $G^\textrm{img}$.

\subsubsection{Gated Multi-modal Fusion.}
Cross-modal fusion of ID and image representations is essential for multi-modal interaction, but direct application of static weighting may induce modality conflicts.
To address this issue, we employ a gated fusion mechanism that dynamically adjusts the contribution of each modality. Specifically, we first concatenate $\mathbf{Z}_p^{\textrm{fusion}} \in \mathbb{R}^{N_p \times d}$ and $\mathbf{Z}_\textrm{img}^{\textrm{fusion}}  \in \mathbb{R}^{N_p \times d}$, then apply a linear transformation followed by a Sigmoid activation to generate gate values $g\in \mathbb{R}^{N_p \times d}$ constrained within [0, 1]; finally, these gate values weight the two modalities:
\begin{equation}
\begin{split}
     g &= \textrm{Sigmoid}(W[\mathbf{Z}_p^\textrm{fusion} \ \Vert \ \mathbf{Z}_\textrm{img}^\textrm{fusion}]+b),\\
     \mathbf{Z}^m &= g\ \odot \ \mathbf{Z}_p^\textrm{fusion} + (1-g)\ \odot \ \mathbf{Z}_\textrm{img}^\textrm{fusion},
\end{split}
\end{equation}
where $\mathbf{W} \in \mathbb{R}^{d \times 2d}$ is a weight matrix, $\mathbf{b} \in \mathbb{R}^{d}$ is a bias vector, $\parallel$ denotes concatenation, and $\odot$ is element-wise multiplication.
We then perform residual weighting between the multi-modal graph representations $\mathbf{Z}^m$ and the original image/ID modality representations respectively, written as:
\begin{equation}
\begin{split}
     \hat{\mathbf{Z}}^p &= \alpha\mathbf{Z}^m + (1-\alpha)\mathbf{Z}^p,\\
     \hat{\mathbf{Z}}^\textrm{img} &= \alpha  \mathbf{Z}^m+ (1-\alpha)\mathbf{Z}^\textrm{img},
\end{split}
\end{equation}
where $\mathbf{Z}^p$ denotes the original one-hot embeddings of POI, $\alpha$ denotes a hyperparameter that controls the proportion of multi-modal graph fused information.
This residual connection can effectively help avoid the degradation of cross-modal representations caused by the intra-modal noise.

\subsubsection{Multi-level User Preference.}
To capture users' preferences for different levels of locations, we construct a multi-level user preference representation following the hierarchical knowledge of
textual semantics, written as:
\begin{equation}
\mathcal{P}_\text{multi}=\mathcal{P}_p +  \mathcal{P}_c + \mathcal{P}_a,
\end{equation}
where the preference $ \mathcal{P}_k $ 
is computed by $\mathcal{P}_k = \textrm{ent}^u + r^t + 
\textrm{ent}^k, k\in{\{p,c,a\}}$, by summing the embeddings of the visit triplet $(\textrm{ent}^u, r^t, \textrm{ent}^k)$ from STKG.
During the training process, we freeze all entity and relation representations to avoid preference forgetting while fine-tuning down-stream prediction tasks.

\subsection{Multi-modal Trajectory Optimization}

\subsubsection{Hierarchical Regularization.}
The mobility sequence representation starts from a visit record of $(u, p, t)$ and its associated category and activity following the hierarchical knowledge of textual semantics, written as:
\begin{equation}
\mathbf{Z}^r = f(\hat{\mathbf{Z}}^p,\mathbf{Z}^u)\ \Vert \ f(\hat{\mathbf{Z}}^c,\hat{\mathbf{Z}}^a,\mathbf{Z}^t) \ \Vert \ \mathcal{P}_\textrm{multi} \ \Vert \  \hat{\mathbf{Z}}^\textrm{img},
\end{equation}
where $\mathbf{Z}^u$ denotes the one-hot encoded representation of user IDs and $f(\cdot)=\textrm{MLP}(\textrm{concat}(\cdot))$. A day is discretized into 48 time slots, each spanning 30 minutes, and temporal representations $\mathbf{Z}^t$ are derived using Time2Vector~\cite{kazemi2019time2vec} to capture temporal periodicity.

The corresponding representation $S_{u_i}$ of the mobility sequence of user $u_i$ is computed by $E_{S_{u_i}}=[\mathbf{Z}^{r_1},\mathbf{Z}^{r_2},\cdots,\mathbf{Z}^{r_{|S_{u_i}|}}]$. 
The obtained representation matrix of the mobility sequence is then fed into the Transformer decoder to obtain the final representation, written as:
\begin{equation}
    \hat{h}_{|S_{u_i}|}= \textrm{Transformer}([\mathbf{Z}_1,\mathbf{Z}_2,\cdots,\mathbf{Z}_{|S_{u_i}|}]),    
\end{equation}
where the Transformer computes the attention scores of the historical locations iteratively as $\textrm{softmax}(\frac{QK^T}{\sqrt{d}})V$.

By incorporating the next category, activity, and time as auxiliary tasks, prediction performance for the next location can be enhanced.
Therefore, we construct four MLP prediction heads to simultaneously perform four prediction tasks, written as:
\begin{equation}
\mathbf{y}_{k}=\textrm{MLP}_k (\hat{h}_{S_{u_i}}),k\in\{p,c,a,t\}.
\end{equation}

\subsubsection{STKG-Guided Cross-Modal Alignment.}
To bridge the semantic gap between static multi-modal data and dynamic mobility behaviors, we enrich the static image representations with the spatial-temporal dynamics from the Hierarchical Spatial-Temporal Knowledge Graph (STKG).
Specifically, we first input the location entity from the pretrained STKG and the image into two independent projection layers and align them in the same space.
\begin{equation}
\begin{split}
\mathbf{Z}^\textrm{KG} = \textrm{Proj}_\textrm{KG}(\textrm{ent}^p),
\end{split}
\end{equation}
where $\textrm{Proj}_\textrm{KG}$ is a projection layer composed of a single layer MLP. 
The alignment loss between image-modal and STKG entity are computed in a global way, which treats the corresponding $\textrm{img}_i$ and $p_i$ as the positive sample and $\textrm{img}_i$ and $p_j (j \neq i)$ as negative samples, written as:
\begin{equation}
\label{eq:contrast}
\begin{split}
\mathcal{L}_{\text{con}} = -\frac{1}{N_p} \sum_{i=1}^{N_p}  
& \log \Bigg[ \frac{\exp\big(\textrm{sim}(\mathbf{Z}_{i}^{\text{KG}}, \mathbf{Z}_{i}^{\text{img}})\big)}{\sum_{j=1}^{N_p} \exp\big(\textrm{sim}(\mathbf{Z}_{i}^{\text{KG}}, \mathbf{Z}_{j}^{\text{img}})\big)}\\
+& \log \frac{\exp\big(\textrm{sim}(\mathbf{Z}_{i}^{\text{img}}, \mathbf{Z}_{i}^{\text{KG}})\big)}{\sum_{j=1}^{N_p} \exp\big(\textrm{sim}(\mathbf{Z}_{i}^{\text{img}}, \mathbf{Z}_{j}^{\text{KG}})\big)}\Bigg],
\end{split}
\end{equation}
where $\textrm{sim}(\cdot)$ denotes the inner product.
Bidirectional contrastive learning can prevent modal collapse. Cross-modal alignment enhances the spatial-temporal dynamics of the image modality.

\subsubsection{Training Objective.}
Finally, we adopt a multi-task learning approach to collaboratively optimize four prediction losses and cross-modal alignment losses to enhance the next location recommendation, our training objective integrates all previous loss functions, written as:
\begin{equation}
    \mathcal{L}_{} =  \mathcal{L}_p + \mathcal{L}_c +\mathcal{L}_a +\lambda_{t}
\mathcal{L}_t+ \lambda_{con}\mathcal{L}_\textrm{con},
\end{equation}
where $\mathcal{L}_p$, $\mathcal{L}_c$ and $\mathcal{L}_a$ denote the cross-entropy losses for next location, next category and next activity prediction, respectively, and $\mathcal{L}_t$ represents the mean squared error loss for next time prediction. $\lambda_{t}$, $\lambda_{con}$ is the weights of $\mathcal{L}_t$, $\mathcal{L}_\textrm{con}$, respectively.

\section{Experiments}
We conduct extensive experiments on real trajectory dataset of six cities to verify the performance of \name in next-location recommendation task. 
We aim to answer the following research questions:
\begin{itemize}[leftmargin=*]
\item \textbf{RQ1}: How does \name perform compared with other state-of-the-art methods for the next-location recommendation task?
\item \textbf{RQ2}: How is the generalization capability of \name under abnormal scenarios like adverse weather and long-tail locations?
\item \textbf{RQ3}: How does \name compare with baselines in efficiency?
\item \textbf{RQ4}: How do different modules affect the prediction performance of \name across various mobility datasets?
\item \textbf{RQ5}: How do hyper-parameter influence \name's performance?
\end{itemize}

\subsection{Experiment Setup}
\begin{table}[t]
    \small
    \caption{Basic statistics of mobility datasets.}
  \begin{tabular}{ccccccc}
    \toprule
    \textbf{Datasets} & \textbf{\#User} & \textbf{\#Loc} &\textbf{\#Record} &\textbf{\#Cat}&\textbf{\#Traj} \\
    \midrule
     NYC & \num{1331} & \num{9312}  & \num{103012}   & 352& \num{22244}\\
     SGP & \num{1997} & \num{8561}  & \num{215885}   & 325& \num{44255}\\
     MEX & \num{2324} & \num{11574} & \num{139312}  & 352& \num{34946}\\
     JKT & \num{3189} & \num{13597} & \num{223802}  & 355& \num{51635}\\
     MOW & \num{3907} & \num{15047} & \num{278362}  & 363& \num{64698}\\
     TKY & \num{3803} & \num{12618} & \num{424535}  & 325& \num{85391}\\
  \bottomrule
\end{tabular}

\label{tab:data}
\end{table}

\begin{figure}[t]
	\centering
	\includegraphics[width=\columnwidth]{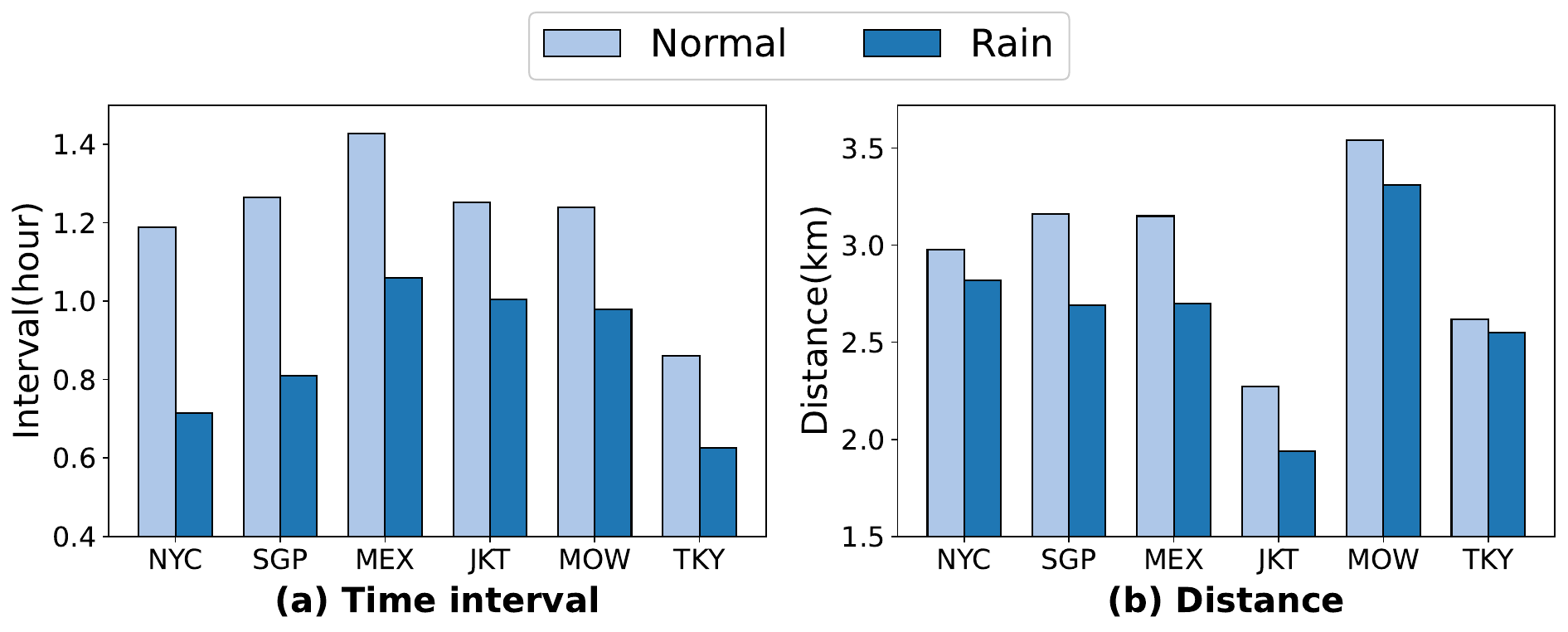} %
	\caption{Urban Variations in Time Intervals and Distances Between Normal and Rainy Weather.}
	\label{fig:rain1}
\end{figure}

\begin{figure}[t]
	\centering
	\includegraphics[width=\columnwidth]{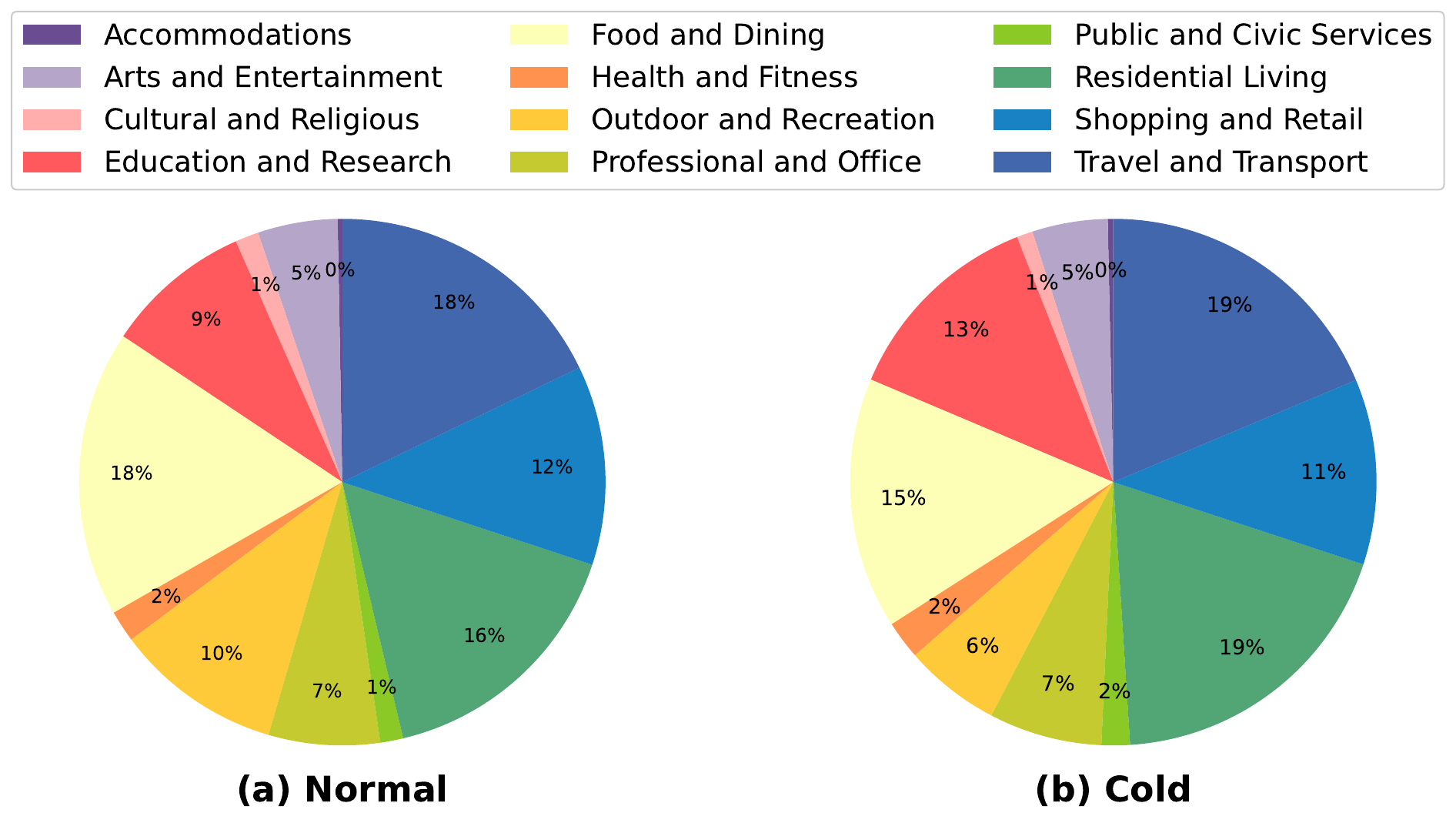} %
	\caption{Human Activity in Normal \textit{VS} Cold Weather.}
	\label{fig:low_temperature.pdf}
\end{figure}
\subsubsection{Datasets.}
To train and evaluate our model's performance, we conduct extensive experiments on trajectory datasets from six cities sourced from Foursquare\footnote{https://sites.google.com/site/yangdingqi/home/foursquare-dataset}. We perform a general preprocessing step~\cite{sun2020go}, filtering locations with fewer than 10 visits; defining all check-ins within a calendar day as a single trajectory; removing trajectories with fewer than 3 check-ins; and excluding inactive users with fewer than 5 trajectories. Each user's trajectory is split chronologically: 80\% training, 10\% validation, and 10\% test. Table~\ref{tab:data} summarizes the statistics of the experimental datasets. Image representations are derived from 256$\times$256-pixel satellite imagery acquired via ArcGIS\footnote{https://server.arcgisonline.com/} at three granularities: coarse (zoom level 15), medium (zoom level 16), and fine (zoom level 17). Weather data are obtained from ERA5-Land\footnote{https://cds.climate.copernicus.eu/datasets}. Figure~\ref{fig:rain1} compares the time intervals and geographical distances between consecutive locations in trajectories under rainy and sunny conditions across six cities. Figure~\ref{fig:low_temperature.pdf} compares the proportion of human activities on MOW under colder periods versus normal weather conditions. 

\begin{table*}[t]

\centering
\small
\caption{Performance comparison against \emph{state-of-the-art} baseline methods. The best results are denoted by \textbf{bold font}, and the second best ones are denoted by \underline{underscore}. Each model was trained five times independently with different random seeds, and the results are reported as mean ± standard deviation.}
\setlength{\tabcolsep}{1.0mm}

\begin{tabular}{lccccccccc}
\toprule
\multirow{2}{*}{Method} &\multicolumn{3}{c}{NYC} &\multicolumn{3}{c}{SGP} &\multicolumn{3}{c}{MEX} \\
\cmidrule(lr){2-4} 
\cmidrule(lr){5-7}
\cmidrule(lr){8-10}
&Acc@5 (\%)&Acc@10 (\%)&Acc@20 (\%)&Acc@5 (\%)&Acc@10 (\%)&Acc@20 (\%)&Acc@5 (\%)&Acc@10 (\%)&Acc@20 (\%)\\
\midrule
Markov     & \num{29.10(0.00)}& \num{33.80(0.00)}& \num{35.71(0.00)}& \num{32.15(0.00)}& \num{39.52(0.00)}& \num{45.24(0.00)}& \num{23.72(0.00)}& \num{27.93(0.00)}& \num{30.58(0.00)} \\ 
LSTM       &\num{33.66(0.21)}& \num{41.04(0.32)}& \num{46.76(0.25)} & \num{39.51(0.34)}& \num{48.45(0.30)}& \num{56.53(0.25)}& \num{32.42(0.23)}& \num{37.94(0.29)}& \num{42.75(0.31)}\\ 
Deepmove    & \num{37.00(0.38)}& \num{44.54(0.41)}& \num{50.50(0.39)}& 
\num{42.30(0.43)}& \num{51.92(0.43)}& \num{60.74(0.32)}& \num{36.30(0.30)}& 
\num{42.70(0.33)}& \num{48.00(0.47)}        \\
LSTPM& \underline{38.54} $\pm$ 0.53& \underline{46.31} $\pm$ 0.31& \num{52.51(0.47)}& \underline{43.22} $\pm$ 0.19& \num{52.91(0.19)}&	\num{61.89(0.15)}& \underline{37.34} $\pm$ 0.21&	\num{44.39(0.26)}& \num{50.13(0.41)}\\
Flashback  & \num{37.37(0.26)}& \num{45.34(0.25)}& \num{51.68(0.21)} & \num{41.97(0.16)}& \num{51.46(0.19)}& \num{60.12(0.14)}& \num{36.33(0.19)}& \num{43.07(0.10)}& \num{48.85(0.11)}\\
MCLP        & \num{36.87(0.19)}& \num{44.51(0.25)}& \num{50.46(0.09)}& \num{42.36(0.06)}& \num{51.26(0.23)}& \num{59.82(0.32)}& \num{36.23(0.29)}& \num{43.16(0.29)}& \num{48.72(0.08)}\\ 
\midrule
GETNEXT     & \num{34.58(0.18)}& \num{42.01(0.30)}& \num{48.34(0.41)}& \num{40.90(0.13)}& \num{48.93(0.13)}& \num{56.35(0.24)}& \num{30.95(0.12)}& \num{36.99(0.23)}& \num{41.98(0.32)}\\
Graph-Flashback &\num{36.84(0.19)}& \num{45.59(0.14)}& \num{52.55(0.14)}& \num{41.85(0.22)}& \num{52.33(0.15)}& \num{61.56(0.07)}& \num{36.97(0.33)}& \num{44.74(0.20)}& \num{50.89(0.14)} \\ 
Diff-POI        & \num{29.22(0.17)}& \num{39.30(0.19)}& 	\num{46.50(0.17)} & \num{29.75(0.12)}& \num{40.98(0.11)}& \num{51.05(0.21)}& \num{29.93(0.56)}& \num{38.45(0.27)}& \num{44.65(0.20)}\\
MMPOI           & \num{36.71(0.24)}& \num{45.86(0.08)}& \underline{52.72} $\pm$ 0.17& \num{42.78(0.15)}& \underline{53.41} $\pm$ 0.16& \underline{62.73} $\pm$ 0.12 &\num{37.06(0.19)}	& \underline{44.86} $\pm$ 0.15& \underline{51.18} $\pm$ 0.19\\
LoTNext     & \num{35.58(0.16)}& \num{43.81(0.07)}& \num{50.65(0.25)}& \num{42.52(0.01)}& \num{52.32(0.12)}& \num{61.07(0.19)}& \num{34.28(0.13)}& \num{42.04(0.28)}& \num{48.40(0.25)}\\
\midrule
\name & \textbf{39.91} $\pm$ 0.17& \textbf{48.59} $\pm$ 0.07& \textbf{54.96} $\pm$ 0.25& \textbf{45.82} $\pm$ 0.13& \textbf{55.99} $\pm$ 0.08& \textbf{64.99} $\pm$ 0.13& \textbf{38.96} $\pm$ 0.16& \textbf{46.56} $\pm$ 0.12& \textbf{52.95} $\pm$ 0.22\\
\midrule
Improv. \% & 3.55 & 4.92 & 4.25& 6.02& 4.83& 3.60& 4.34& 3.79& 3.46\\
\midrule

\multirow{2}{*}{Method} &\multicolumn{3}{c}{JKT} &\multicolumn{3}{c}{MOW} &\multicolumn{3}{c}{TKY} \\
\cmidrule(lr){2-4} 
\cmidrule(lr){5-7}
\cmidrule(lr){8-10}
&Acc@5 (\%)&Acc@10 (\%)&Acc@20 (\%)&Acc@5 (\%)&Acc@10 (\%)&Acc@20 (\%)&Acc@5 (\%)&Acc@10 (\%)&Acc@20 (\%)\\
\midrule
Markov      & \num{25.23(0.00)}& \num{30.82(0.00)}& \num{35.61(0.00)}& \num{23.50(0.00)}& \num{28.60(0.00)}& \num{32.60(0.00)}& \num{29.97(0.00)}& \num{37.65(0.00)}& \num{44.51(0.00)}\\ 
LSTM        & \num{31.63(0.22)}& \num{39.53(0.23)}& \num{47.74(0.17)}& \num{31.96(0.20)}& \num{38.06(0.29)}& \num{43.38(0.36)}& \num{36.84(0.24)}& \num{44.07(0.45)}& \num{51.05(0.62)}\\ 
Deepmove    & \num{35.56(0.40)}& \num{44.30(0.47)}& \num{53.40(0.40)}& \num{36.46(0.43)}& \num{43.78(0.66)}& \num{49.98(0.72)}& \num{41.34(0.63)}& \num{49.98(0.54)}& \num{57.77(0.60)}\\
LSTPM       & \underline{36.46} $\pm$ 0.17& \num{45.36(0.17)}& \num{54.25(0.24)}& \num{37.17(0.37)}& \num{45.18(0.27)}& \num{51.84(0.29)}& \underline{43.75} $\pm$ 0.10& \num{52.18(0.13)}& \num{59.61(0.11)}\\
Flashback   & \num{35.36(0.07)}& \num{44.03(0.19)}& \num{52.90(0.12)}& \num{36.33(0.11)}& \num{43.07(0.11)}& \num{48.85(0.14)}& \num{39.35(0.12)}& \num{48.16(0.11)}& \num{56.03(0.05)}\\
MCLP        & \num{35.02(0.09)}& \num{43.38(0.10)}& \num{51.96(0.23)}& \num{36.42(0.14)}& \num{44.24(0.25)}& \num{50.63(0.19)}& \num{40.96(0.07)}& \num{49.29(0.13)}& \num{56.84(0.13)}\\ 
\midrule
GETNEXT     & \num{32.10(0.27)}& \num{39.32(0.34)}& \num{46.93(0.55)}& \num{33.55(0.13)}& \num{40.35(0.28)}& \num{46.13(0.26)}& \num{38.91(0.30)}& \num{47.14(0.31)}& \num{54.70(0.25)}\\
Graph-Flashback & \num{36.39(0.09)}& \underline{46.06} $\pm$ 0.18& \underline{55.42} $\pm$ 0.15& \num{37.21(0.19)}& \num{45.44(0.12)}& \num{52.42(0.17)}& \num{40.50(0.13)}& \num{49.95(0.11)}& \num{58.11(0.08)}\\ 
Diff-POI        & \num{27.95(0.19)}& \num{36.85(0.09)}& \num{45.55(0.11)}& \num{31.33(0.32)}& \num{40.33(0.26)}& \num{47.98(0.18)}& \num{29.87(0.09)}& \num{39.76(0.09)}& \num{48.47(0.05)}\\
MMPOI      & \num{35.74(0.22)}& \num{45.53(0.20)}& \num{54.82(0.21)}& \underline{38.02} $\pm$ 0.17& \underline{46.60} $\pm$ 0.14& \underline{53.63} $\pm$ 0.20& \num{42.66(0.12)}&	\underline{52.38} $\pm$ 0.11)& \underline{60.69} $\pm$ 0.11\\
LoTNext         & \num{34.85(0.07)}& \num{44.56(0.12)}& \num{54.20(0.08)}& \num{35.70(0.12)}& \num{44.31(0.14)}& \num{51.51(0.15)}& \num{40.79(0.08)}& \num{50.01(0.05)}& \num{58.46(0.08)}\\
\midrule
\name & \textbf{37.61} $\pm$ 0.11& \textbf{47.53} $\pm$ 0.22& \textbf{57.31} $\pm$ 0.20& \textbf{39.25} $\pm$ 0.09& \textbf{47.91} $\pm$ 0.12& \textbf{55.11} $\pm$ 0.13& \textbf{44.61} $\pm$ 0.15& \textbf{53.76} $\pm$ 0.09& \textbf{62.12} $\pm$ 0.10\\
\midrule
Improv. \% & 3.15& 3.19&  3.41& 3.24& 2.81& 2.76& 1.97& 2.63& 2.36\\
\bottomrule
\end{tabular}

\label{tab:comparison}
\end{table*}


\subsubsection{Baselines.}
To validate \name's effectiveness, we benchmark it against 11 state-of-the-art baselines. These comparative baselines are categorized into sequence-based methods~\cite{gambs2012next,hochreiter1997long,feng2018deepmove,yang2020location,sun2020go,sun2024going} and graph-based methods~\cite{yang2022getnext,rao2022graph,qin2023diffusion,xu2024mmpoi,xu2024taming}. More details are provided in Appendix B.1.

\subsubsection{Evaluation Metrics.} We evaluate predictions using $\textrm{Acc}@k$, the standard location recommendation metric. This metric assesses whether the true next location appears in the top-$k$ recommendations, effectively measuring system accuracy. 

\subsection{RQ1: Performance Comparison}

Table~\ref{tab:comparison} shows the performance of different baselines on 6 datasets. Our method consistently outperforms state-of-the-art baselines across all datasets, achieving average improvements of 3.71\% at Acc@5, 3.70\% at Acc@10, and 3.31\% at Acc@20. Among sequence-based methods, LSTPM excels by explicitly modeling long- and short-term preferences, while MCLP struggles with sparse trajectory data and topic modeling fails to capture diverse user preferences. Graph-based methods like GETNext and Diff-POI show no improvement over sequence approaches: GETNext introduces noise via its Transition Attention Map, and Diff-POI underutilizes temporal sequences. Although Graph-Flashback and LoTNext leverage spatial-temporal knowledge graphs, they neglect real-world multi-modal information. MMPOI leads other graph-based methods but primarily uses multi-modal data for intra-modal similarity graphs and fails to leverage high-quality multi-modal explicit representations. See Appendix B.3 for hyperparameter search results of the latest baseline methods (MCLP, LoTNext).

\begin{figure}[t]
	\centering
	\includegraphics[width=\columnwidth]{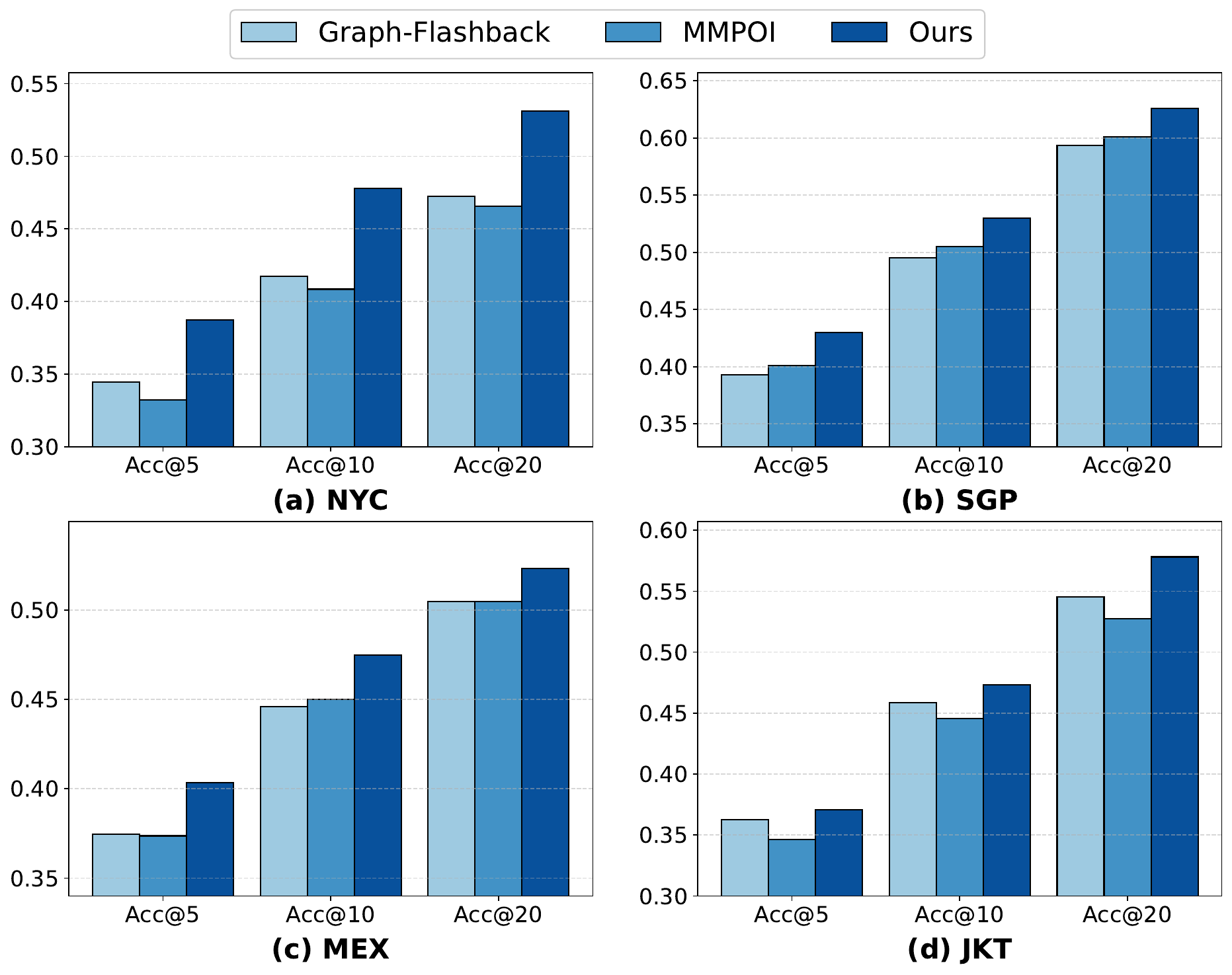}
	\caption{Prediction accuracy on rainy weather.}
	\label{fig:rain_prediction}
\end{figure}

\begin{figure}[t]
	\centering
	\includegraphics[width=\columnwidth]{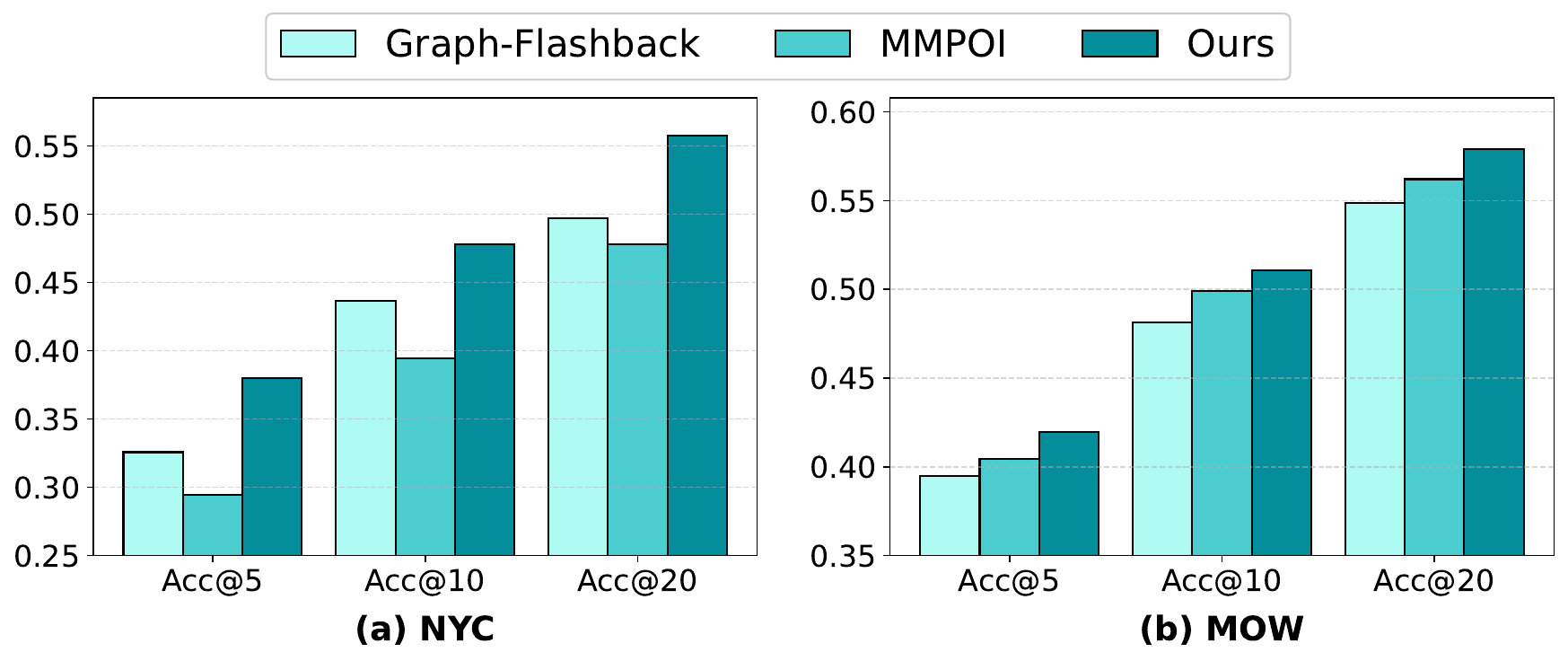} %
	\caption{Prediction accuracy on cold weather.}
	\label{fig:cold_prediction}
\end{figure}

\begin{figure}[t]
	\centering
	\includegraphics[width=\columnwidth]{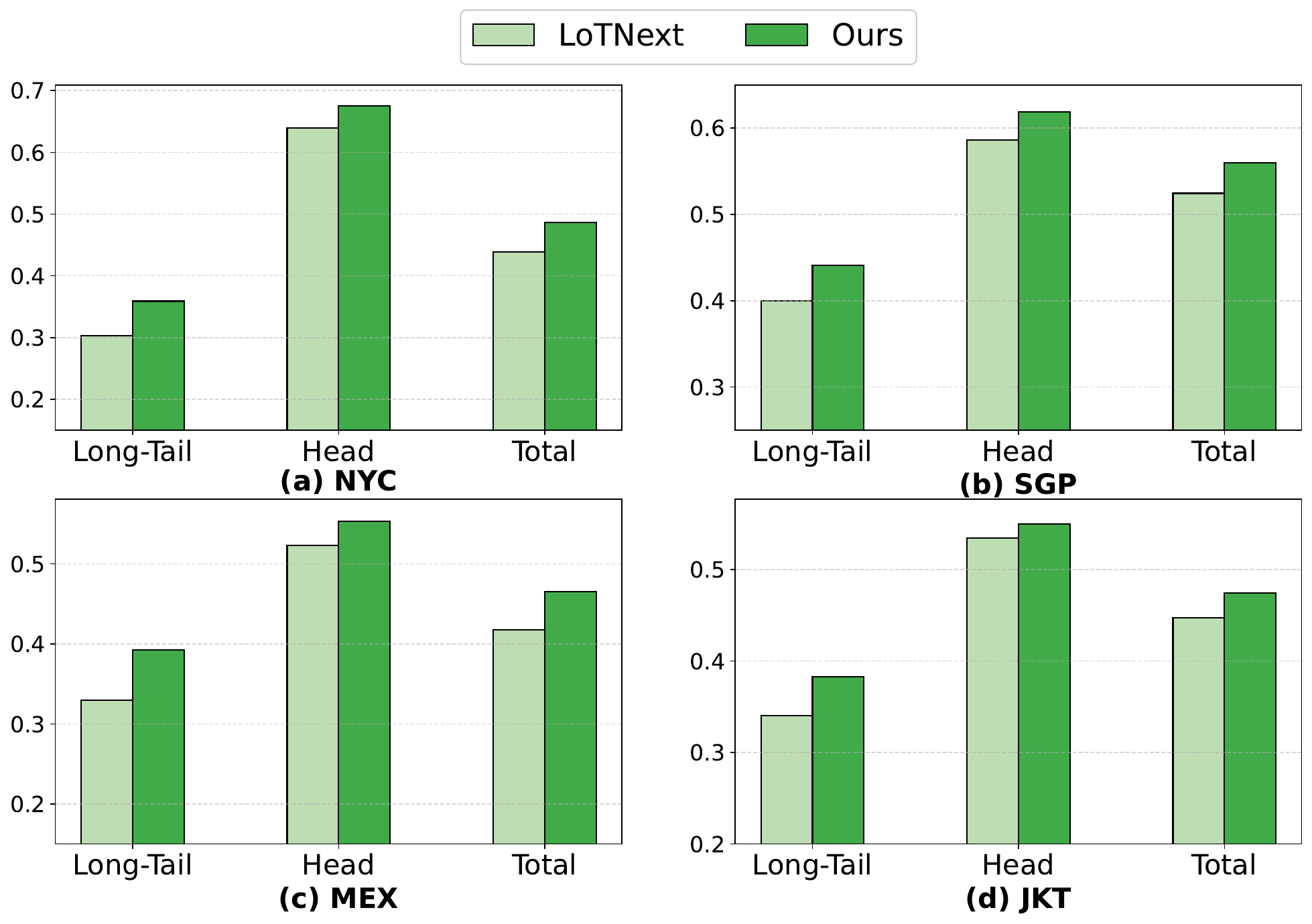} %
	\caption{Prediction accuracy on long-tail locations. The performance is measured using Acc@10.}
	\label{fig:longtail}
\end{figure}
\subsection{RQ2: Generalization Analysis of \name}
\subsubsection{Performance Analysis on Adverse Weather.}
Human travel preferences shift significantly in rainy and cold weather~\cite{cools2010changes,horanont2013weather,brum2018weather}. As shown in Figure~\ref{fig:rain1}, analysis of trajectory data from six cities shows that during rain, average time intervals and spatial distances decreased by 28.2\% and 9.7\% respectively compared to sunny conditions. Movement speed increased, indicating a preference for visiting nearer locations and greater use of vehicles. NYC and MOW, situated at higher latitudes, are frequently affected by cold spells. As shown in Figure~\ref{fig:low_temperature.pdf}, in cold weather, people favor indoor activities, reducing trips for dining and shopping. Such specific weather conditions challenge the generalization of mobility models in sparse scenarios.
As shown in Figure~\ref{fig:rain_prediction} and Figure~\ref{fig:cold_prediction}, our method significantly outperforms Graph-Flashback and MMPOI in location recommendation under both rainy and cold weather. This consistent advantage across datasets confirms our approach’s generalization in adverse weather. This is because the remote sensing image module in our method enriches the environmental semantics of locations, while the text semantics of LLM captures high-level differences in human activities. Additionally, the unified STRG and the STKG-guided cross-modal alignment enhance the spatiotemporal knowledge across modalities, enabling the capture of dynamic shifts in human behavior under adverse weather conditions.


\subsubsection{Performance Analysis on Long-tail Locations.}
Long-tail locations (frequency $< 20$) pose significant challenges in mobility prediction due to the lack of collaborative signals. Our proposed \name addresses this by leveraging frequency-agnostic multi-modal data. As shown in Figure~\ref{fig:longtail}, our method consistently outperforms LoTNext (a long-tail adjustment framework) in predicting both head and tail locations. The performance of LoTNext is limited by the sparse trajectory data and inherent data biases, whereas our approach enriches information for long-tail locations by leveraging real-world textual and visual modalities. Moreover, through a shared spatial-temporal relational graph, our method endows different modalities with dynamic spatial-temporal representations, enabling effective semantic transfer from head locations to tail ones.

\begin{table}[t]
\centering
\small
\caption{Model efficiency comparison.}
\setlength{\tabcolsep}{1mm}
\begin{tabular}{lcrccrc}
\toprule
\multirow{2}{*}{Method} & \multicolumn{3}{c}{NYC} & \multicolumn{3}{c}{SGP} \\ 
\cmidrule(lr){2-7}
& Memory & Train & Infer & Memory & Train & Infer  \\
\midrule
Deepmove            & 1,141M& 1.20& 1.27 & 1,579M& 2.45& 2.50\\
LSTPM               & 1,781M& 23.42& 3.20& 1,609M& 47.80& 5.88\\
Graph-Flashback     & 3,913M& 0.52& 0.83& 3,783M& 1.20& 1.53\\
LoTNext             & 2,211M& 2.35& 1.08& 1,323M& 5.13& 2.22 \\
MMPOI               & 5,263M& 1.53& 1.25& 5,519M& 3.43& 2.40\\
\name               & 4,931M& 0.43& 0.33& 5,055M& 1.02& 0.80\\ 
\hline
\multicolumn{7}{l}{\footnotesize Note: M (megabyte),  Train/Infer (minute).} \\

\end{tabular}

\label{tab:efficiency}
\end{table}

\subsection{RQ3: Efficiency Evaluation}
As shown in Table~\ref{tab:efficiency}, our method demonstrates superior training and inference efficiency over competitive baselines on NYC and SGP datasets: Among sequence-based methods, LSTPM incurs substantially higher training/validation time by serially processing each historical trajectory; for graph-based methods, LoTNext requires additional time overhead per iteration due to graph denoising operations. Furthermore, despite utilizing multi-modal data, our lightweight model construction introduces little additional memory overhead compared to unimodal methods.

\begin{table}[t]
\centering
\small
\setlength{\tabcolsep}{0.95mm}
\caption{Ablation analysis of \name. The recommendation
performance is measured using Acc@10 (\%).}
\begin{tabular}{lcccc}
\toprule
\multirow{1}{*}{Variants} & \multicolumn{1}{c}{NYC} & \multicolumn{1}{c}{SGP} & \multicolumn{1}{c}{MEX} & \multicolumn{1}{c}{JKT} \\ 

\midrule
\multicolumn{5}{c}{Multi-modal Data}\\
\midrule
w/o Img & \num{46.29(0.13)}& \num{53.94(0.09)}& \num{43.92(0.33)}& \num{45.51(0.21)}\\ 
w/o Text& \num{48.16(0.11)}& \num{55.61(0.07)}& \num{46.19(0.25)}& \num{47.12(0.26)}\\ 
\midrule
\multicolumn{5}{c}{Multi-modal Fusion}\\
\midrule
w/o IRG  & \num{48.30(0.16)}& \num{55.66(0.24)}& \num{46.24(0.13)}& \num{47.28(0.15)} \\
w/o STRG & \num{48.02(0.11)}& \num{55.35(0.11)}& \num{45.69(0.10)}& \num{47.02(0.28)}\\
w/o MUP  & \num{48.31(0.13)}& \num{55.69(0.16)}&  \num{46.15(0.19)}& \num{47.29(0.20)}    \\
\midrule
\multicolumn{5}{c}{Multi-modal Alignment}\\
\midrule
w/o CMA  & \num{47.87(0.18)}& \num{55.65(0.26)}& \num{46.08(0.08)}& \num{47.16(0.10)}\\
\midrule
\name  & $ \textbf{48.59}\,\pm\,0.07 $ & $ \textbf{55.99}\,\pm\,0.08 $ & $ \textbf{46.56}\,\pm\,0.12 $ & $ \textbf{47.53}\,\pm\,0.22 $      \\ 
\bottomrule
\end{tabular}
\label{tab:ablation}
\end{table}
\subsection{RQ4: Ablation Study}
To evaluate the effectiveness of different modules, we constructed six variants of \name in our ablation study. As shown in Table~\ref{tab:ablation}, the results demonstrate the importance of our multi-modal integration and alignment designs. The variants are analyzed as follows:
\begin{itemize}[leftmargin=*]
    \item w/o Img: Removing the remote sensing image modality and cross-modal alignment leads to a clear performance drop (-4.58\%), as images are capable of effectively enriching the spatial hierarchy and environmental semantic information of locations.
    \item w/o Text: Excluding high-level textual semantics from the knowledge graph and related loss functions reduces performance, since text helps cluster locations with similar categories and enriches hierarchical semantic representation.
    \item w/o IRG: Removing the image relation graph (IRG), which links mobility dynamic behavior to static images, causes a decline by widening the semantic gap between behavior and static image.
    \item w/o STRG: When the spatial-temporal relational graph (STRG) is excluded, the IRG is also removed simultaneously—this is because the IRG is derived from the STRG. The STRG is capable of modeling dynamic relationships between multiple modalities and mitigating the sparsity issue; consequently, the model’s performance decreases (-1.31\%) after this module is removed.
    \item w/o MUP: Omitting multi-level user preferences (MUP), which capture time-aware user interests across location semantics, results in observable performance degradation.
    \item w/o CMA: Eliminating the cross-modal alignment (CMA), which infuses images with spatial-temporal dynamics from the knowledge graph, also leads to a performance drop.
\end{itemize}



\subsection{RQ5: Hyper-parameter Sensitivity}
\begin{figure}[t]
	\centering
	\includegraphics[width=\columnwidth]{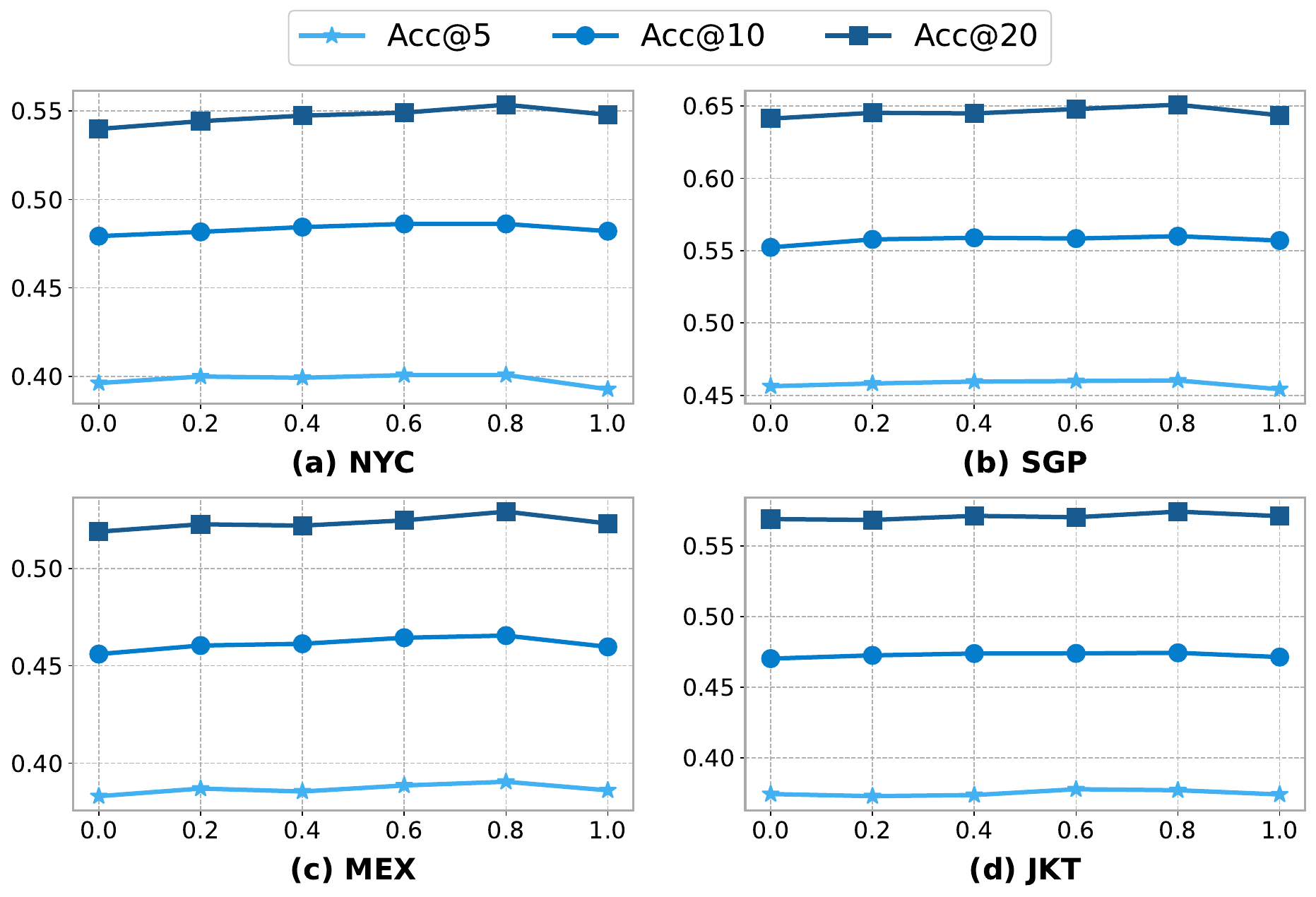} %
	\caption{Effect of the graph fusion weight $\alpha$.}
	\label{fig:fusion}
\end{figure}

Figure~\ref{fig:fusion} illustrates variation in prediction accuracy of \name with the multi-modal representation fusion weight $\alpha$. Our hyperparameter search over $\alpha$ in the range [0.0, 0.2, 0.4, 0.6, 0.8, 1.0] reveals that performance significantly degrades without multi-modal graph representations ($\alpha$ = 0), underscoring critical importance of spatial-temporal graph fusion; notably, omitting multi-modal residual connections ($\alpha$ = 1) leads to a marked performance decline on four datasets, indicating these residuals effectively mitigate information loss across modalities. Furthermore, a fusion weight of $\alpha$ = 0.8 consistently yields near-optimal results across all datasets. Hyperparameter sensitivity experiments for K-NN on STRG and multitask loss weights can be found in Appendix B.4.

\section{Conclusion}
In this study, we introduce \name, a multi-modal fusion framework that leverages spatial-temporal knowledge to integrate human mobility dynamics with multi-modal representations, build a united STRG for multi-modal and implements STKG-guided cross-modal contrastive alignment.
Our method exhibits strong generalization ability, as demonstrated by extensive comparative experiments conducted across both normal and abnormal scenarios—especially under adverse weather and long-tail locations. Benefiting from its lightweight design, the method also boasts high inference efficiency. Furthermore, ablation studies validate the contributions of different modules within our framework.
In the future, we will explore dynamic scenarios for multi-modal mobility prediction to enhance the emergency management capabilities of decision-makers.
Another promising direction is to investigate the interpretable human mobility paradigm by guiding the mobility of LLMs agents with their perceptual multi-modal knowledge.

\clearpage
\bibliographystyle{ACM-Reference-Format}
\bibliography{main}

\clearpage
\appendix

\section{Method Details.}

\subsection{LLM Prompt for Location Hierarchy.}

\begin{figure}[H]
	\centering
	\includegraphics[width=\columnwidth]{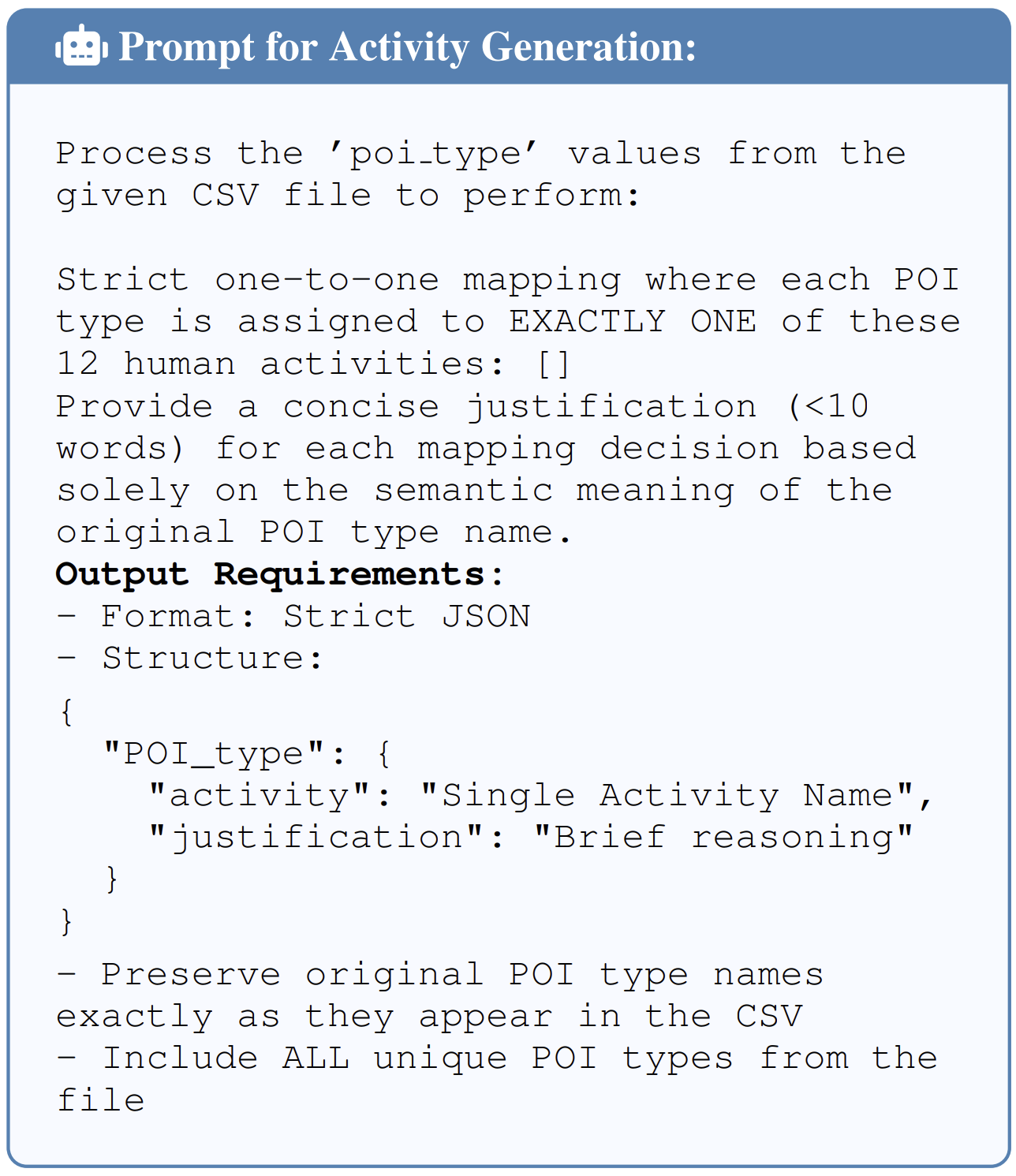} %
	\caption{Category $\rightarrow$ Activity Prompt.}
	\label{fig:prompt}
\end{figure}

The prompt used for generating Activity labels from Category data is shown as Figure~\ref{fig:prompt}. Based on the categorical text descriptions of locations, we employ a large language model to infer 12 generic types of human activities.

\section{Experiments.}

\subsection{Baselines.}

To validate the effectiveness of \name, we benchmark it against 11 state-of-the-art baselines. These comparative baselines are categorized into sequence-based methods and graph-based methods.
\begin{parenum}[leftmargin=*]
\item Sequence-based model
\begin{itemize}[leftmargin=*]
\item  Markov~\cite{gambs2012next}:
It regards all visited locales as states, and constructs a transition matrix to capture the first-order transition probabilities between them.

\item  LSTM~\cite{hochreiter1997long}: LSTM is a sophisticated variant of RNN, designed to adeptly manage sequential data.

\item  Deepmove~\cite{feng2018deepmove}: It employs the attention mechanism to more effectively leverage historical data and contextual information during prediction.

\item LSTPM~\cite{sun2020go}: It proposes using non-local networks and geographically dilated RNNs to model long- and short-term preferences.
\item  Flashback~\cite{yang2020location}: It employs the flashback mechanism to address historical trajectories.

\item  MCLP~\cite{sun2024going}: It employs thematic modeling to extract users' historical locational preferences and generates arrival time embeddings as the context for location recommendation through a multi-head attention mechanism.
\end{itemize}

\item Graph-based model
\begin{itemize}

\item  GETNEXT~\cite{yang2022getnext}: It employs trajectory flow graph to capture the general movement patterns of users.
\item  Graph-Flashback~\cite{rao2022graph}: Based on the spatial-temporal knowledge graph, enhance location representation by utilizing the temporal and spatial relationships between locations.
\item  Diff-POI~\cite{qin2023diffusion}: It harnesses two graphs to extract spatial-temporal representations, and introduces a diffusion-based sampling strategy to investigate users' spatial preferences.
\item MMPOI~\cite{xu2024mmpoi}: It leverages intra-modal cosine similarity to construct a multi-modal graph, thereby enriching location representation.
\item LoTNext~\cite{xu2024taming}: It proposes long-tailed graph adjustment and a long-tailed loss adjustment module to enhance long-tailed location prediction.

\end{itemize}
\end{parenum}

\subsection{Implementation Details.}
All experiments are performed on a single NVIDIA RTX 4090. We utilize Adam as the optimizer, with an initial learning rate set to 1e-4 and the L2 regularization penalty set to 1e-3. For the image encoder, we adopt CLIP's pre-trained image encoder (ViT-B-32). For Transformer, we stack two layers of encoders with a dropout of 0.3 and set the number of attention heads to 4 to run 75 epochs with batch size 128. We search for ID dimension from $\{128, 256, 512\}$. It is worth noting that we set the time loss and alignment loss weights to 10 and 1 respectively to match the scale of location and category losses. Additionally, we set the weights for the graph fusion representation to 0.8. For a fair comparison, we optimized each baseline's parameters as per their papers and recorded the mean and standard deviation across five random seeds.

\begin{table}[t]
\centering
\small
\caption{Hyperparameter grid search results for MCLP and LoTNext on random seed 42.}
\setlength{\tabcolsep}{1mm}
\begin{tabular}{cccccccccc}
\toprule
MCLP & \multicolumn{3}{c}{NYC} & \multicolumn{3}{c}{SGP}\\
{[batch, lr, dim]}&Acc@5 & Acc@10 & Acc@20&Acc@5 & Acc@10 & Acc@20\\ 
\midrule
{[64,1e-4,128]}& 0.3675& 0.4423& 0.5012& 0.4142& 0.5073& 0.5930\\
{[64,1e-4,256]}& 0.3701& \textbf{0.4462}& \textbf{0.5079}& 0.4224& 0.5124& 0.5976\\
{[64,1e-3,128]}& 0.3488& 0.4261& 0.4890& 0.4053& 0.4957& 0.5837\\
{[64,1e-3,256]}& 0.3433& 0.4204& 0.4873& 0.4041& 0.4980& 0.5818\\
{[128,1e-4,128]}& 0.3647& 0.4392& 0.5029& 0.4152& 0.5070& 0.5906\\
{[128,1e-4,256]}& 0.3591& 0.4314& 0.4841& 0.4225& 0.5127& 0.5976\\
{[128,1e-3,128]}& 0.3574& 0.4316& 0.4911& 0.4041& 0.4984& 0.5833\\
{[128,1e-3,256]}& 0.3528& 0.4291& 0.4935& 0.4006& 0.4931& 0.5770\\
{[256,1e-4,128]}& 0.3659& 0.4405& 0.4975& 0.4163& 0.5078& 0.5920\\
{[256,1e-4,256]}& \textbf{0.3705}& 0.4458& 0.5070& \textbf{0.4244}& \textbf{0.5159}& \textbf{0.6019}\\
{[256,1e-3,128]}& 0.3553& 0.4350& 0.4907& 0.4067& 0.5016& 0.5840\\
{[256,1e-3,256]}& 0.3565& 0.4352& 0.4986& 0.4010& 0.4946& 0.5774\\
\midrule
LoTNext & \multicolumn{3}{c}{NYC} & \multicolumn{3}{c}{SGP}\\
{[batch, lr, dim]}&Acc@5 & Acc@10 & Acc@20&Acc@5 & Acc@10 & Acc@20\\ 
\midrule
{[16,1e-4,128]}& 0.3549& 0.4229& 0.4796& 0.4198& 0.5118& 0.5974\\
{[16,1e-4,256]}& 0.3565& 0.4338& 0.4900& 0.4273& 0.5210 & 0.6062\\
{[16,1e-3,128]}& 0.3560& 0.4383& 0.5083& 0.4147& 0.5158& 0.6054\\
{[16,1e-3,256]}& 0.3529& \textbf{0.4386}& \textbf{0.5106}& 0.4133& 0.5163& 0.6118\\
{[32,1e-4,128]}& 0.3516& 0.4241& 0.4760& 0.4215& 0.5178& 0.6025\\
{[32,1e-4,256]}& \textbf{0.3608}& 0.4348& 0.4903& 0.4228& 0.5195& 0.6058\\
{[32,1e-3,128]}& 0.3575& 0.4364& 0.4986& 0.4261& \textbf{0.5244}& \textbf{0.6132}\\
{[32,1e-3,256]}& 0.3490& 0.4283& 0.4963& 0.4224& 0.5218& 0.6089\\
{[128,1e-4,128]}& 0.3517& 0.4202& 0.4765& 0.4207& 0.5186& 0.6028\\
{[128,1e-4,256]}& 0.3591& 0.4314& 0.4841& \textbf{0.4268}& 0.5233& 0.6078\\
{[128,1e-3,128]}& 0.3421& 0.4137& 0.4774& 0.4161& 0.5083& 0.5918\\
{[128,1e-3,256]}& 0.3528& 0.4291& 0.4935& 0.4219& 0.5187& 0.6059\\

\bottomrule
\end{tabular}

\label{tab:Appendix_grid_MCLP}
\end{table}


\subsection{Hyperparameter Search of Baselines.}


Given their status as the latest sequence-based model (MCLP) and graph-based model (LoTNext) among our baseline comparisons, we conducted comprehensive grid hyperparameter search experiments for both methods. The results of these searches for selected parameters on the NYC and SGP datasets are presented in Table~\ref{tab:Appendix_grid_MCLP}.

Despite extensive hyperparameter tuning, we observe that both methods continue to underperform prior work (e.g., LSTPM and Graph-Flashback). This performance gap occurs because MCLP was evaluated on datasets with lower sparsity than ours; their approach performs best with denser trajectory data. Furthermore, we identified a data leakage issue in the publicly released LoTNext code, where the Transformer module lacked temporal masking during inference. We confirmed this flaw via email correspondence with the authors and subsequently added masking to LoTNext in our experiments to ensure fair comparison.
\subsection{More Parameter Analysis.}
\begin{figure}[t]
	\centering
	\includegraphics[width=\columnwidth]{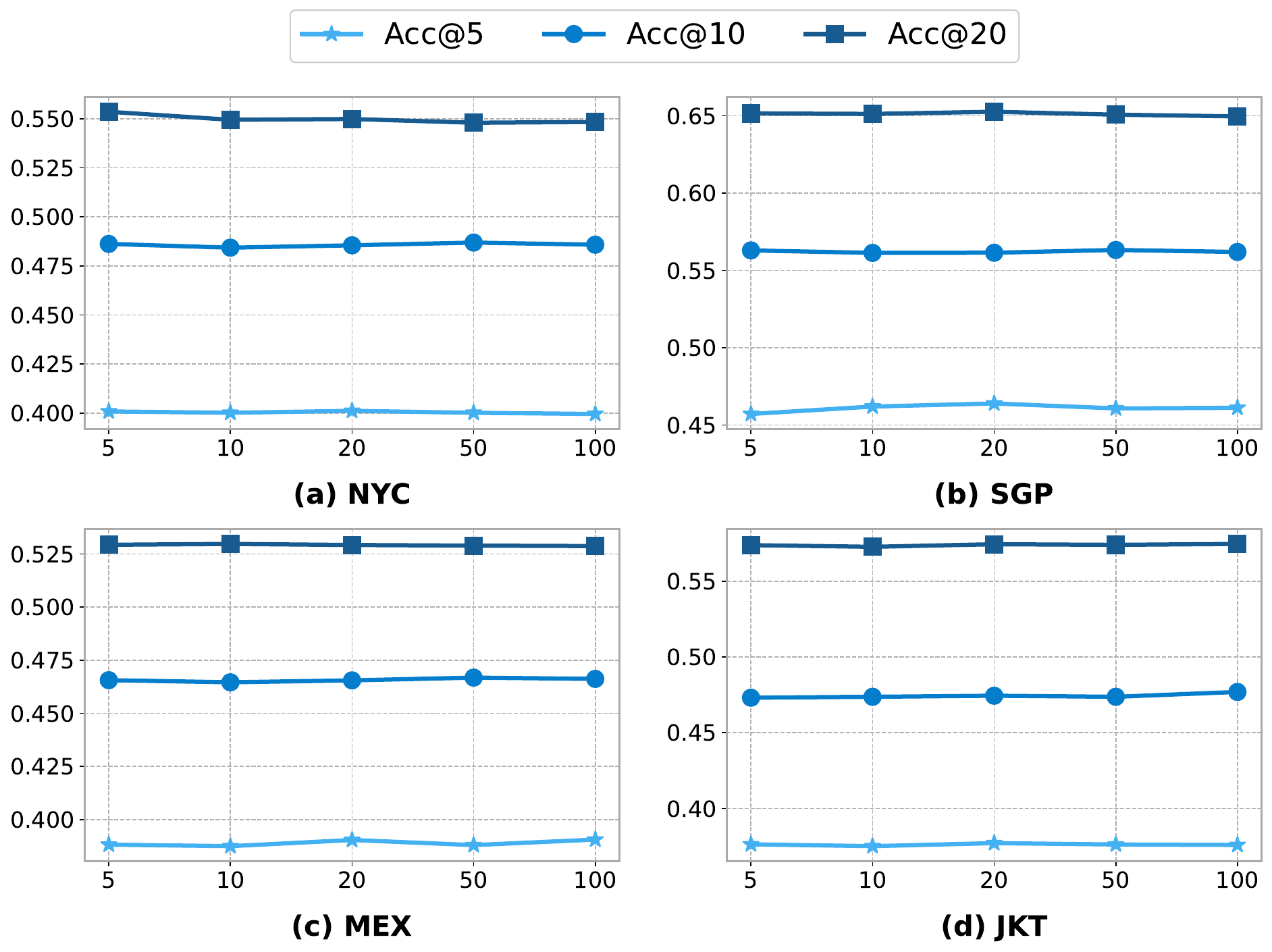} %
	\caption{Effect of the $k$ of $k$-NN for spatial-temporal Relational Graph.}
	\label{fig:KNN}
\end{figure}

\begin{figure}[t]
	\centering
	\includegraphics[width=\columnwidth]{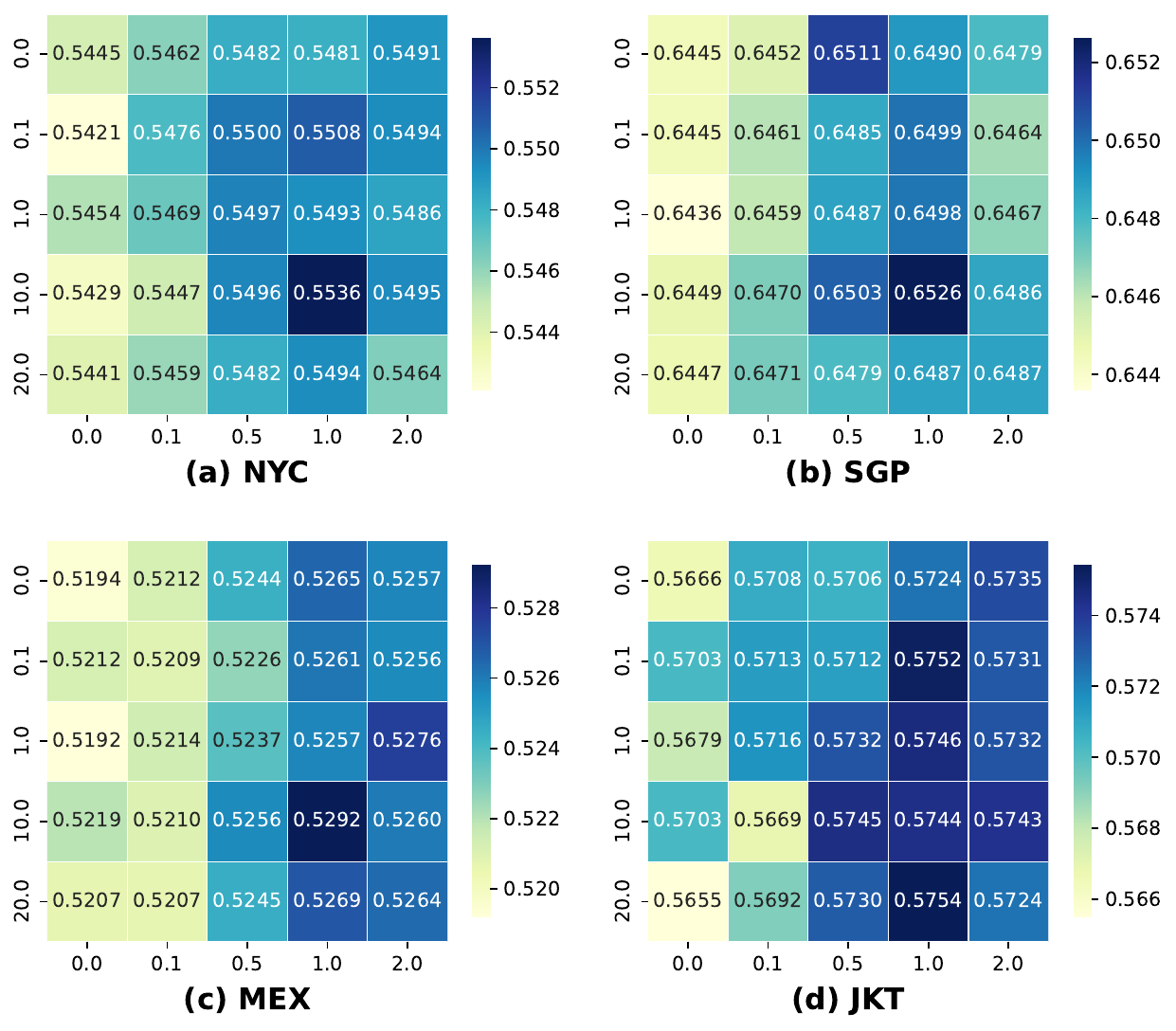} %
	\caption{Effect of the weights of multitask loss. The performance is measured using Acc@20.}
	\label{fig:loss_weight}
\end{figure}
\subsubsection{K-NN for Spatial-Temporal Relational Graph.} Figure~\ref{fig:KNN} demonstrates the prediction accuracy variation of \name with respect to the neighborhood size parameter \(k\) in the KNN-sparsified graph. Our hyperparameter search over \(k \in \{5, 10, 20, 50, 100\}\) reveals distinct patterns: For SGP, MEX and JKT, accuracy initially increases with larger \(k\) values and saturates at \(k=20\) as incorporating more spatial-temporally relevant neighbors enriches multi-modal representations of locations; beyond this point, performance slightly declines due to noise introduced by weakly correlated neighbors. Conversely, for NYC, optimal accuracy is achieved at \(k=5\) with no significant improvement at larger values, which stems from extreme data sparsity—characterized by low average visit frequencies and abundant long-tail locations possessing few strongly associated neighbors.

\subsubsection{Multitask Loss Weights.}
To analyze the impact of multi-task loss weights on the performance of our method, we searched for the alignment loss weight $\lambda_\textrm{con}$ within the range [0, 0.1, 0.5, 1, 2] and the temporal prediction loss weight $\lambda_t$ within [0, 0.1, 1, 10, 20]. Figure~\ref{fig:loss_weight} shows how the performance of our method varies with changes in the multi-task loss weights, where the x-axis represents $\lambda_\textrm{con}$ and the y-axis represents $\lambda_t$. We observe that removing either the temporal loss or the alignment loss leads to a degradation in the performance of our method. Additionally, excessively high weights for both losses will cause an imbalance in the multi-task loss, resulting in performance degradation. Across the four datasets, $\lambda_t = 10$ and $\lambda_\textrm{con} = 1$ are found to be favorable values.
\subsection{Impact of Scaling Ratio on Remote Sensing.}
\begin{figure}[t]
	\centering
	\includegraphics[width=\columnwidth]{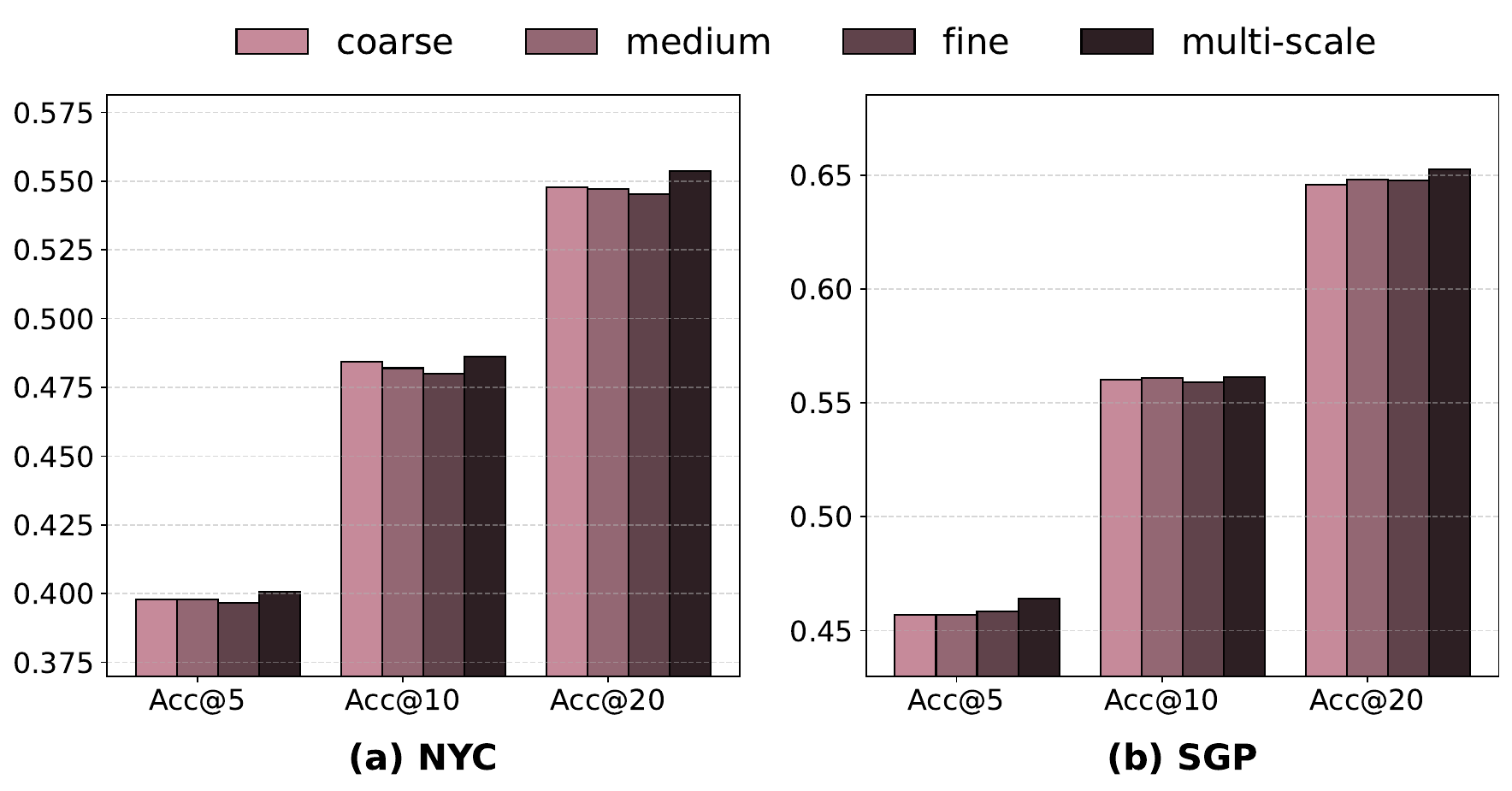} %
	\caption{Effect of scaling ratio on remote sensing.}
	\label{fig:remote_scale}
\end{figure}
Remote sensing images with different scaling ratios exhibit distinct spatial structural semantics: high-scaling-ratio images contain more fine-grained information, while low-scaling-ratio ones have a larger receptive field. As shown in Figure~\ref{fig:remote_scale}, we compared the location recommendation performance of two cities with distinct morphological characteristics, using coarse-, medium-, fine-grained remote sensing images, and the multi-scale remote sensing images adopted in our framework as image modal representations respectively.
We found that for NYC, a city with concentrated functional zones and large spatial scales, coarse-grained remote sensing images can capture macro functional zone features and achieve the best performance. For SGP, a city with mixed functions and compact spaces, medium/fine-grained remote sensing images are more effective as they capture microscopic neighborhood features. Meanwhile, the multi-scale remote sensing image representation outperforms those with a single granularity, as it integrates information from remote sensing images of various granularities.
\end{document}